\documentclass[10pt,conference,letterpaper]{IEEEtran}
\IEEEoverridecommandlockouts

\usepackage{amsmath,amssymb,amsthm}
\usepackage{siunitx}

\usepackage{graphicx}
\usepackage{caption}
\usepackage{subcaption}
\usepackage{float}
\usepackage{booktabs}
\usepackage{multirow}
\usepackage{tabularx}
\usepackage{colortbl}
\usepackage{xcolor}
\usepackage{textcomp}
\usepackage{afterpage}
\usepackage{gensymb}

\usepackage{enumitem}
\usepackage{array}
\newcolumntype{L}[1]{>{\raggedright\arraybackslash}m{#1}}
\newcolumntype{C}[1]{>{\centering\arraybackslash}m{#1}}
\newcolumntype{R}[1]{>{\raggedleft\arraybackslash}m{#1}}

\usepackage{algorithm}
\usepackage[noend]{algpseudocode}  

\algrenewcommand\algorithmicrequire{\textbf{Require:}}
\algrenewcommand\algorithmicensure{\textbf{Ensure:}}

\usepackage{hyperref}
\usepackage{cite}

\newcommand{\boldtheorem}[1]{\textbf{#1}}
\newtheorem{theorem}{\boldtheorem{Theorem}}
\newtheorem{lemma}{\boldtheorem{Lemma}}

\newtheorem{assumption}{\boldtheorem{Assumption}}

\def\BibTeX{{\rm B\kern-.05em{\sc i\kern-.025em b}\kern-.08em
  T\kern-.1667em\lower.7ex\hbox{E}\kern-.125emX}}

\title{
MetaMolGen: A Neural Graph Motif Generation Model for De Novo Molecular Design
\thanks{
This work was supported by the National Natural Science Foundation of China [61773020] and the Graduate Innovation Project of National University of Defense Technology [XJQY2024065]. The authors would like to express their sincere gratitude to all the referees for their careful reading and insightful suggestions.
}
}

\author{%
{ Zimo Yan\textsuperscript{\rm 1}, Jie Zhang\textsuperscript{\rm 1}, Zheng Xie\textsuperscript{\rm 1}\thanks{*Corresponding author: Zheng Xie (xiezheng81@nudt.edu.cn).}, Chang Liu\textsuperscript{\rm 1}, Yizhen Liu\textsuperscript{\rm 1}, Yiping Song\textsuperscript{\rm 1} }%
\vspace{1.6mm}\\
\fontsize{10}{10}\selectfont\itshape
\textsuperscript{\rm 1}National University of Defense Technology, Changsha, China.
\\\{yanzimo20, zhangjie, xiezheng81, nudt\_liuchang\_1997, liuyizhen23, songyiping\}@nudt.edu.cn}
\fontsize{9}{9}\selectfont\ttfamily\upshape
\fontsize{10}{10}\selectfont\rmfamily\itshape

\fontsize{9}{9}\selectfont\ttfamily\upshape

\begin{document}
\maketitle

\begin{abstract}
Molecular generation plays an important role in drug discovery and materials science, especially in data-scarce scenarios where traditional generative models often struggle to achieve satisfactory conditional generalization. To address this challenge, we propose MetaMolGen, a first-order meta-learning-based molecular generator designed for few-shot and property-conditioned molecular generation. 
MetaMolGen standardizes the distribution of graph motifs by mapping them to a normalized latent space, and employs a lightweight autoregressive sequence model to generate SMILES sequences that faithfully reflect the underlying molecular structure. In addition, it supports conditional generation of molecules with target properties through a learnable property projector integrated into the generative process.
Experimental results demonstrate that MetaMolGen consistently generates valid and diverse SMILES sequences under low-data regimes, outperforming conventional baselines. This highlights its advantage in fast adaptation and efficient conditional generation for practical molecular design.

\end{abstract}

\section{Introduction}

Designing and synthesizing molecules with specific properties is a fundamental challenge in fields like drug discovery, materials science, and environmental chemistry \cite{Shoichet, Scior}. The vastness of chemical space makes molecular design inherently complex. While over $10^8$ molecules have been synthesized, this represents only a tiny fraction of the estimated $10^{23}$ to $10^{60}$ potential drug-like molecules \cite{Poli, Reymond}, highlighting both the immense potential and the difficulty of molecular exploration.

When employing intelligent models for molecular design, ensuring structural validity remains a critical challenge. Current molecular generation models primarily adopt one of three strategies. \textbf{Graph-based methods} generate molecules atom-by-atom or bond-by-bond, enabling high interpretability and direct manipulation of molecular topology; a representative example is StrucGCN~\cite{zhang2025strucgcn}, which improves graph embedding by incorporating structural information. \textbf{Sequence-based models} utilize linear representations such as SMILES, and although they offer strong scalability, they often struggle with generating valid structures. Approaches like RNNs~\cite{Segler} and Transformer-based models~\cite{Bagal2022} fall into this category. \textbf{Tree-based methods}, such as Junction Tree VAE~\cite{Jin W}, generate molecules by assembling subgraph fragments in a hierarchical manner, achieving 100\% structural validity.

Beyond these mainstream paradigms, emerging strategies are addressing more specific challenges in molecular generation. Diffusion-based models, like PIDiff~\cite{Choi2024}, leverage iterative denoising to construct 3D molecular conformations, while retrieval-augmented methods such as Reinvent~\cite{Schneider} enhance synthetic feasibility and diversity by integrating database search with generative models. Recent advancements including MGCVAE~\cite{Lee M} and GAN-based frameworks like ORGAN~\cite{Guimaraes2017} further improve multi-property control and support task-specific molecular generation.

Despite these advancements, existing methods still suffer from persistent limitations that restrict their effectiveness in practical applications. A key challenge lies in the difficulty of simultaneously ensuring both validity and uniqueness, making it problematic to generate chemically diverse yet valid molecules tailored for specific properties. For instance, MolGAN~\cite{De Cao} achieves nearly perfect validity (e.g., 98\%--100\%) on MOSES benchmarks, but exhibits low uniqueness and diversity (often below 10\%), revealing a clear trade-off. To provide a comprehensive evaluation of model performance across multiple objectives, we define an \textit{Overall Score} that integrates validity, uniqueness, diversity, and druglikeness. The formulation of the Overall Score is given in Equation~\ref{eq:overall_score}.

The data-intensive nature of deep learning approaches presents another significant barrier, typically demanding training sets on the order of $10^6$ samples. This requirement becomes prohibitive in data-scarce scenarios such as orphan drug discovery, severely limiting their applicability to smaller datasets. For example, sequence-based methods like GrammarVAE~\cite{Kusner2017GrammarVAE} generally require vast amounts of data to achieve optimal performance, whereas our meta-learning-based approach can effectively leverage prior knowledge to rapidly adapt with only a fraction of the data, yielding superior results in few-shot settings.

\begin{figure}[h]
	\centering
	\includegraphics[width=0.48\textwidth]{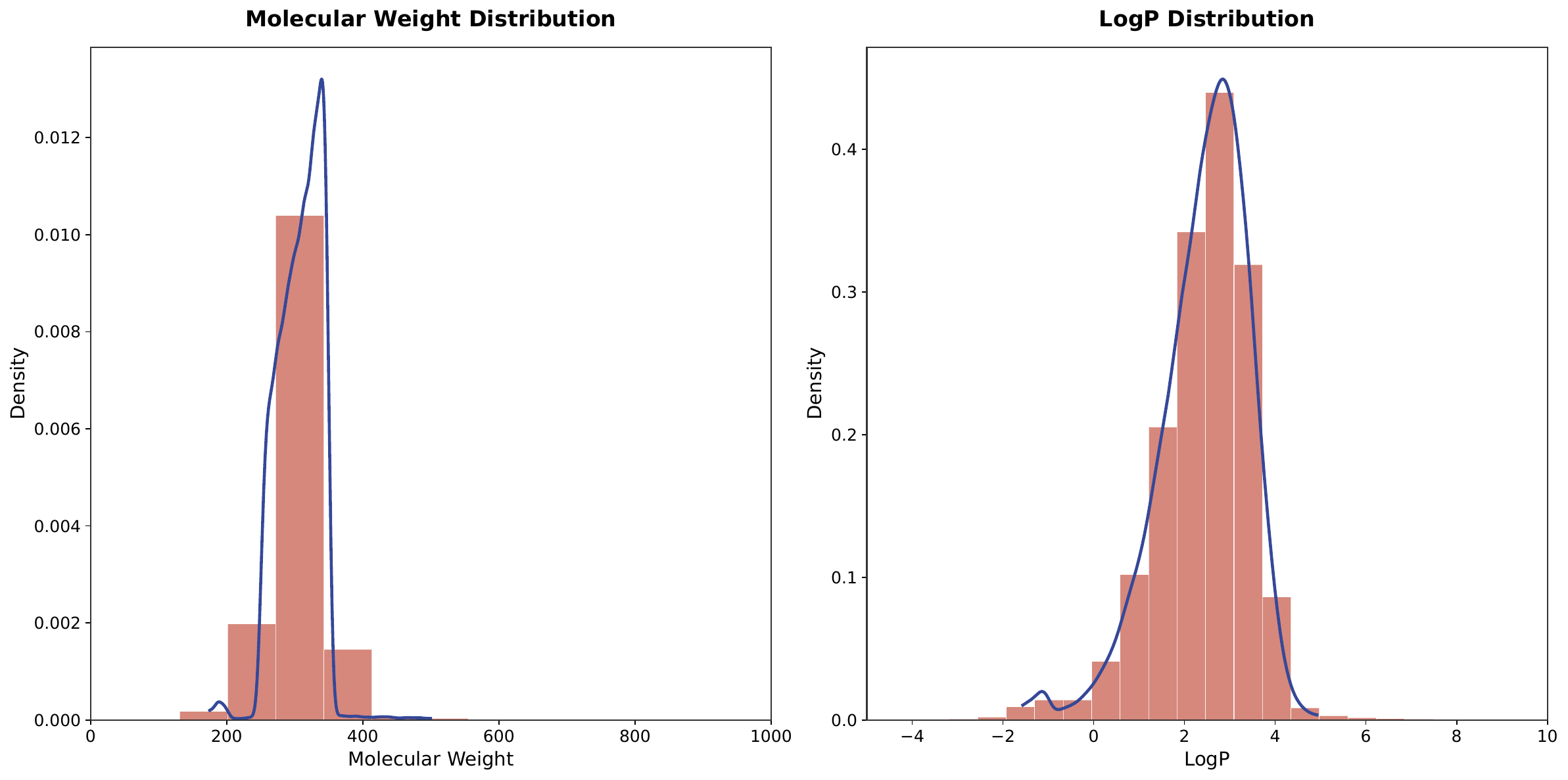}
	\caption{Distribution of Key Molecular Properties}
	\label{datadistribution}
\end{figure}

More critically, while many molecular generation models require complete datasets to fully capture the nuances of molecular properties, our preliminary analysis reveals that some key properties, particularly LogP, tend to exhibit approximately unimodal distributions (Fig.~\ref{datadistribution}). In contrast, molecular weight is right-skewed and hydrogen bond acceptor and donor counts display more complex patterns; yet even these can largely be seen as following a unimodal trend. This predominance of unimodal characteristics in continuous properties like LogP suggests a potential statistical bias in molecular datasets that may be underutilized. Notably, this bias implies that our approach could tolerate partial data or even benefit from using only a fraction of the available data, thereby alleviating the reliance on large-scale, complete datasets. Further investigation is warranted to determine how leveraging this bias might improve generation performance.

\textbf{Motivation}:
This leads to the following question: \textit{Can we design a model that not only generates molecules under small-sample conditions but also uses the unimodal property distribution to enhance and balance the model's validity and uniqueness?}

To address this challenge, we designed the Meta-Optimized Molecular Generation model (MetaMolGen), which leverages feature transformation and small-sample learning techniques to overcome these limitations. The MetaMolGen framework focuses on generating valid molecules that maintain diversity while explicitly modeling unimodal distributions in the chemical space.

\textbf{Developed Methods}:
We developed MetaMolGen, a deep learning model based on the Conditional Neural Processes (CNPs) framework, which overcomes the limitations of traditional point estimation methods by learning function spaces rather than single-point functions, thereby capturing broader feature variations. This innovation effectively addresses the inherent shortcomings of the MolGAN approach, achieving a significant improvement in molecular uniqueness while sacrificing only a small portion of validity (approximately 5-10\%).

To address data scarcity conditions, we integrated the Reptile meta-learning algorithm \cite{nichol2018first} into the CNPs framework. Reptile enables efficient adaptation to small-scale datasets (typically between $10^4$ and $10^5$ samples)  by learning a robust initialization through first-order updates across tasks, without requiring second-order gradient computation. Notably, for the observed unimodal distribution phenomenon in molecular features, we introduced a normalization layer prior to the context encoder, transforming the feature distribution into a quasi-normal form.

\textbf{Outline \& Primary Contributions}:  
This paper is organized as follows. Section~\ref{SE2} provides a review of related work in molecular generative models and meta-learning.  
Section~\ref{SE3} introduces the necessary preliminaries, notations, and problem formulation.  
Section~\ref{SE4} presents the proposed MetaMolGen method, including architectural design, feature normalization, and training framework.  
Section~\ref{SE5} offers theoretical analysis covering assumptions, convergence guarantees, and generalization bounds.  
Section~\ref{SE6} describes the experimental setup, including datasets, baselines, and evaluation metrics.  
Section~\ref{SE7} reports empirical results and compares model performance across tasks.  
Section~\ref{SE8} provides an in-depth discussion of model behavior and ablation results.  
Section~\ref{SE9} concludes the paper and outlines potential directions for future work.

The primary contributions of this work are as follows:
\begin{enumerate}
     \item To address the challenge of balancing model validity and uniqueness, we propose MetaMolGen, a novel meta-learning molecular generation model designed for few-shot molecular generation. Additionally, MetaMolGen is capable of conditional generation, enabling the generation of molecules with specific chemical properties based on given target attributes.

    \item Through large-scale experiments on multiple datasets (such as ChEMBL, QM9, ZINC, and MOSES), we validated the advantages of MetaMolGen in molecular generation tasks on small to medium-sized datasets. The model significantly outperforms traditional generative models in terms of validity, uniqueness, property matching accuracy, and generation quality, as reflected in its superior \textit{Overall Score} across key metrics. 

    \item We introduce a learnable standardization module within the CNP framework, which mitigates scale imbalances across molecular features, improving the model's numerical stability and convergence speed.
    \item We integrate Reptile into the CNP framework to achieve rapid adaptation to small-scale datasets, and provide a theoretical discussion on the convergence characteristics of Reptile under this setting.

\end{enumerate}

\section{Literature Review}
\label{SE2}
Recent advances in deep learning have significantly expanded the scope of molecular generative modeling, leveraging architectures such as variational autoencoders (VAEs), generative adversarial networks (GANs), and reinforcement learning (RL)-based methods. These techniques have been applied to a range of tasks, including drug discovery, molecular optimization, and materials design. Despite their success, challenges remain in navigating the high-dimensional molecular space, capturing complex dependencies, and achieving generalization in low-data scenarios.

\subsection{Graph-based Methods}
Graph-based methods represent molecules as undirected graphs, where atoms correspond to nodes and chemical bonds to edges. This structure-aware representation naturally aligns with molecular topology, enabling models to reason over local and global structural information. Graph Neural Networks (GNNs) have been widely used in this context, providing a mechanism for learning rich molecular embeddings.

One early approach, MolGAN, combines GANs with reinforcement learning to generate molecular graphs directly, bypassing the need for SMILES decoding. While MolGAN achieves near-perfect validity on benchmark datasets, it struggles with uniqueness and diversity, achieving less than 10\% uniqueness in some cases. You et al.~\cite{You} proposed the Graph Convolutional Policy Network (GCPN), which combines GNNs with RL to generate molecules that satisfy property constraints, offering a goal-directed design capability.

Further advancements, such as StrucGCN, incorporate structural priors into graph embedding processes to improve model expressiveness. Meanwhile, the Crystal Graph Convolutional Neural Network (CGCNN)~\cite{Xie} has demonstrated that graph-based architectures are also effective in predicting material properties. Despite these advances, graph-based models often require large training datasets to learn meaningful patterns and may struggle to generalize in data-scarce settings.

\subsection{Sequence-based Methods}
Sequence-based models represent molecules as one-dimensional strings, typically using the SMILES notation~\cite{weininger}, allowing generative tasks to be treated similarly to natural language processing. These methods are generally efficient and easy to scale but often suffer from lower validity due to the strict syntactic constraints of SMILES strings.

Recurrent Neural Networks (RNNs) have been widely applied to SMILES generation, learning to produce valid molecules through sequence modeling. Transformer-based models have improved scalability and attention over long sequences, while Conditional VAEs~\cite{Lim J} allow for property-guided generation by conditioning the latent space on molecular attributes. GrammarVAE further incorporates context-free grammar rules into the decoding process, improving validity by enforcing syntactic correctness.

GAN-based methods like ORGAN apply adversarial training and reinforcement learning to SMILES generation, optimizing for target-specific properties via reward signals. Similarly, Reinvent~\cite{Schneider} blends SMILES-based generation with retrieval mechanisms to improve synthetic feasibility and diversity. Despite their flexibility and speed, sequence-based methods tend to have difficulty modeling molecular graphs' complex structural dependencies and often require post-processing to ensure chemical validity.

\subsection{Tree-based Methods}
Tree-based approaches decompose molecules into hierarchies of chemical substructures or motifs and model their generation as an assembly of these components. This paradigm aims to guarantee the syntactic and chemical validity of generated molecules by working at the subgraph level.

The Junction Tree Variational Autoencoder (JT-VAE) represents molecules as junction trees, where each node is a chemical motif. By combining tree-structured generation with graph decoding, JT-VAE guarantees 100\% validity while also enabling structured property optimization. An extension~\cite{Jin1} incorporates conditional generation by conditioning motif assembly on target properties, supporting controllable molecular design.

Tree-based models are particularly effective in generating synthetically feasible and chemically realistic compounds. However, they also introduce complexity in training and inference due to the need for substructure extraction, motif vocabulary construction, and hierarchical decoding. Additionally, they may lack flexibility in discovering entirely novel molecular motifs outside the predefined set.

\subsection{Emerging Directions and Meta-learning for Molecular Generation}

Beyond the three dominant paradigms, newer approaches are emerging to address specific challenges in molecular generation. Diffusion models, such as PIDiff, generate 3D molecular structures through iterative denoising, proving especially valuable for tasks like protein-ligand docking. Retrieval-based generation methods integrate database search with generative modeling to ensure synthetic accessibility and real-world relevance.

In parallel, methods focusing on multi-property control and task-specific compound design have gained increasing attention. MGCVAE enables multi-objective inverse design using conditional graph VAEs, while deep generative frameworks such as those proposed by Zhavoronkov et al.~\cite{Zhavoronkov} have shown notable success in discovering target-specific compounds.

Among these emerging directions, meta-learning has become a particularly promising framework for few-shot molecular generation, especially in data-scarce scenarios such as orphan drug discovery. First-order meta-learning algorithms like Reptile offer a computationally efficient alternative to gradient-based methods such as MAML~\cite{finn2017model}, by optimizing task-agnostic initializations through repeated adaptation across tasks without requiring second-order derivatives. Such approaches enable fast generalization using only a handful of examples.

In the molecular domain, meta-learning is gaining momentum as a solution to the data inefficiency of conventional deep models. Hospedales et al.~\cite{Hospedales} highlight its potential in structured prediction problems, although many existing works still lack explicit integration with molecular graph or motif-based representations.

To this end, our proposed framework, MetaMolGen, adopts a meta-learning strategy tailored to few-shot molecular generation. By combining structural priors with task-adaptive initialization through Reptile, it improves generalization, stability, and sample efficiency in low-resource molecular design scenarios, bridging the gap between data scarcity and effective generative modeling.

\section{Preliminaries}
\label{SE3}
\subsection{Notations}

We consider a molecular dataset \( \mathcal{D} = \{(x_i, y_i)\}_{i=1}^{N} \), where each \( x_i \in \mathbb{R}^d \) denotes a molecular feature vector and \( y_i \) is the corresponding SMILES string (see Section~\ref{subsec:MolRepre} for details). The molecular generation process is formulated as learning a mapping function \( f_\theta: \mathbb{R}^d \to \mathcal{Y} \), where \( \mathcal{Y} \) denotes the space of valid molecular sequences over a vocabulary \( \Sigma = \{\texttt{C}, \texttt{N}, \texttt{O}, \texttt{=}, \ldots\} \). Each element in \( \mathcal{Y} \) corresponds to a valid tokenized SMILES string, where tokens represent atoms, bonds, or structural symbols.

In meta-learning settings, we assume a distribution over tasks \( \mathcal{P}(\mathcal{T}) \), where each task \( \mathcal{T}_m \sim \mathcal{P}(\mathcal{T}) \) consists of a support set \( \mathcal{D}_m^{\text{support}} \) and a query set \( \mathcal{D}_m^{\text{query}} \). The meta-learning objective is denoted by \( \mathcal{L}_{\text{meta}} \), and the model parameters are updated from initial parameters \( \theta \) to task-specific parameters \( \theta_m' \) via a few adaptation steps.

\subsection{Problem Statement}

We consider a conditional molecular generation task, where the goal is to generate syntactically valid and property-compliant SMILES sequences given molecular features and target property constraints.

Formally, given a dataset 
\[
\mathcal{D} = \{(x_i, z_i, y_i)\}_{i=1}^{N},
\]
where \( x_i \in \mathbb{R}^d \) denotes the molecular feature vector, \( z_i \in \mathbb{R}^k \) represents the desired property vector, and \( y_i \in \mathcal{Y} \subset \Sigma^* \) is a tokenized SMILES sequence, the objective is to learn a conditional generative function:
$
f_\theta: (x, z) \mapsto \hat{y}
$
that generates sequences \( \hat{y} \) that are both chemically valid and aligned with the desired properties.

To ensure numerical stability and consistent feature representation, molecular descriptors are standardized using a dataset-dependent affine transformation:
\begin{equation}
X' = \Sigma^{-1/2}(X - \mu), \quad \text{so that} \quad X' \sim \mathcal{N}(0, I)
\label{eq:standardize}
\end{equation}
where the empirical mean and variance are estimated as:
\begin{equation}
\hat{\mu} = \frac{1}{N} \sum_{i=1}^{N} X_i, \quad 
\hat{\sigma} = \sqrt{\frac{1}{N} \sum_{i=1}^{N} (X_i - \hat{\mu})^2}
\label{eq:mean_var}
\end{equation}

During training, the model is optimized via a token-level reconstruction loss to maximize the likelihood of producing ground-truth sequences:
\begin{equation}
\mathcal{L}_{\text{recon}} = \text{CrossEntropy}(\hat{y}, y)
\label{eq:recon}
\end{equation}

For fast task adaptation, the model employs a first-order gradient-based update strategy. For each task \( \mathcal{T}_m \sim p(\mathcal{T}) \), model parameters are adapted by applying a small number of gradient steps on the support set. A single-step update is expressed as:
\begin{equation}
\theta_m' = \theta - \alpha \nabla_\theta \mathcal{L}(\mathcal{D}_m^{\text{support}}; \theta),
\label{eq:inner_update}
\end{equation}
where \( \alpha > 0 \) is the inner-loop learning rate. These task-specific parameters \( \theta_m' \) are then used to compute the Reptile-style meta-update:
\begin{equation}
\theta \leftarrow \theta + \epsilon \cdot \frac{1}{M} \sum_{m=1}^M (\theta_m' - \theta),
\label{eq:reptile_update}
\end{equation}
where \( \epsilon \) is the outer-loop step size. This update rule moves the initialization towards parameters that perform well after a small number of task-specific updates, enabling fast adaptation without requiring second-order gradients. This adaptation rule defines the core mechanism underlying task-level generalization in MetaMolGen, and serves as the basis for our theoretical discussion in Section~\ref{SE5}.

At inference time, molecular sequences are generated autoregressively by modeling the conditional distribution:
\begin{equation}
P_\theta(y \mid x, z) = \prod_{t=1}^{T} P_\theta(y_t \mid y_{<t}, x, z)
\label{eq:autoregressive}
\end{equation}
which ensures syntactic \textit{validity} by learning token transitions compliant with SMILES grammar. To promote output \textit{diversity}, tokens are sampled from a temperature-controlled softmax distribution:
\begin{equation}
y_t \sim \text{Categorical}\left(\text{softmax}\left(\frac{f_\theta(y_{<t}, x, z)}{\tau}\right)\right)
\label{eq:temp_sampling}
\end{equation}

These components jointly support the generation of valid, diverse, and property-aligned molecular structures under limited supervision.

\section{Method}
\label{SE4}

MetaMolGen is a meta-learned molecular generator designed for few-shot molecular generation and multi-objective optimization. Unlike traditional generative models, it leverages the Reptile algorithm~\cite{nichol2018first} for rapid task adaptation, Conditional Neural Processes (CNPs) for task-aware molecular representations, and feature standardization for stable training.

The framework processes molecular data through a structured pipeline. Molecular descriptors are first standardized for numerical stability (Eq.~\eqref{eq:standardize},~\eqref{eq:mean_var}) and then encoded into a task embedding via a context encoder. To enhance structural diversity, MetaMolGen leverages Conditional Neural Processes (CNPs) to capture task-level uncertainty through latent representations conditioned on support sets, where the resulting predictive distribution is formalized as \( p(y \mid x, \mathcal{C}) = \int p(y \mid x, z) \, q(z \mid \mathcal{C}) \, dz \), with \( z \) sampled from the context-dependent posterior \( q(z\mid \mathcal{C}) \).

The learning process in MetaMolGen is based on a meta-learning formulation composed of task-specific adaptation and meta-level optimization.
For fast task adaptation, the model employs a first-order gradient-based update strategy. For each task \( \mathcal{T}_m \sim p(\mathcal{T}) \), model parameters are adapted using the support set through one or more gradient steps, as defined in Eq.~\eqref{eq:inner_update}, where \( \alpha > 0 \) is the inner-loop learning rate. The resulting task-adapted parameters \( \theta_m' \) serve as intermediate optima that guide the subsequent meta-update.
The Reptile algorithm updates the shared initialization \( \theta \) by averaging the differences between the adapted and original parameters across tasks, as shown in Eq.~\eqref{eq:reptile_update}, where \( \epsilon \) is the outer-loop step size and \( M \) denotes the number of sampled tasks. This first-order update enables fast adaptation to novel molecular generation tasks, without relying on second-order derivatives or explicit backpropagation through the adaptation process. It forms the foundation for efficient few-shot generalization in MetaMolGen, as further elaborated in Section~\ref{SE5}.

Sequence generation is performed by an autoregressive decoder, which factorizes the conditional distribution over output tokens as in Eq.~\eqref{eq:autoregressive}. This formulation enforces syntactic validity by modeling token transitions aligned with SMILES grammar. To encourage diverse outputs, decoding is performed via temperature-controlled sampling (Eq.~\eqref{eq:temp_sampling}).

By integrating these components into a coherent pipeline, MetaMolGen effectively balances syntactic validity, structural diversity, and generalization performance, making it well-suited for applications in drug discovery, materials science, and beyond.

\begin{figure}[htbp]
	\centering
	\includegraphics[width=\linewidth]{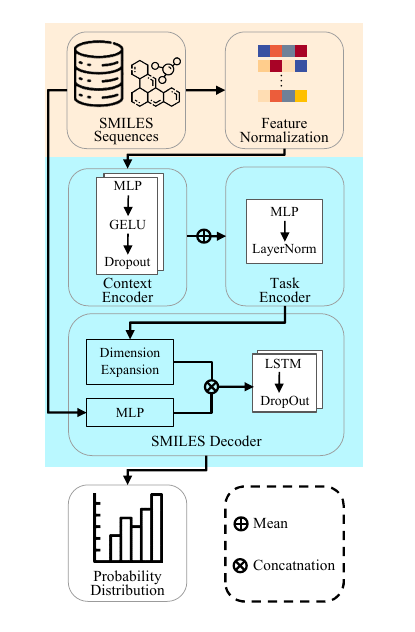}
	\caption{Overview of the MetaMolGen training process.}
	\label{training}
\end{figure}

\subsection{Feature Standardization for Molecular Representation Learning}

To improve training stability and generalization in few-shot molecular generation, MetaMolGen incorporates a learnable normalization layer that standardizes molecular descriptors before task encoding. For each input feature \( x_i \), the model applies an affine transformation:

\begin{equation}
x_i' = \frac{x_i - \mu_\theta}{\sigma_\theta + \varepsilon},
\label{eq:learned_norm}
\end{equation}
where \( \mu_\theta \) and \( \sigma_\theta \) are trainable feature-wise statistics, and \( \varepsilon \) is a small constant to ensure numerical stability. This transformation mitigates input scale disparities and ensures consistent feature distributions across tasks.

Feature normalization plays a critical role in stabilizing optimization and enhancing generalization in few-shot molecular generation. By applying a learnable affine transformation to standardize input features (Eq.~\eqref{eq:learned_norm}), the model mitigates scale imbalance and ensures consistent input distributions across tasks. This reduces sensitivity to outlier features and smooths the optimization landscape, effectively suppressing gradient oscillations during meta-updates, as analyzed in Theorem~\ref{TH2}. A visual comparison of training loss trajectories with and without feature standardization is provided in Appendix C, Figure~\ref{fig:meta_loss_comparison}.

Normalization further improves the conditioning of the loss surface, enabling faster and more stable convergence (Theorem~\ref{TH3}). On the generalization side, by reducing the empirical loss variance across tasks, it tightens the PAC-Bayes generalization bound (Theorem~\ref{TH4}) and helps the model capture task-invariant patterns. Together, these effects make feature normalization a fundamental component of MetaMolGen, supporting both learning stability and transferability under distributional shifts.

\subsection{Molecular Representation}\label{subsec:MolRepre}

This study adopts the SMILES (Simplified Molecular Input Line Entry System) format to represent molecular structures as linear strings. SMILES encodes molecular graphs using a depth-first traversal, translating atoms, bonds, branches, and rings into a sequence of standardized characters. For example, the molecule ethanol is represented as “CCO,” where two carbon atoms are followed by a hydroxyl group. This compact format preserves the topological structure of molecules and is naturally compatible with sequence-based generative models.

In MetaMolGen, SMILES sequences are used as the target output of the decoder. We construct a vocabulary containing special tokens such as \texttt{\{PAD\}}, \texttt{\{START\}}, \texttt{\{EOS\}}, and \texttt{\{UNK\}}, and tokenize each SMILES string into a sequence of discrete tokens. These tokenized sequences are embedded and fed into an LSTM-based decoder, which autoregressively predicts the next token given the previously generated ones. During training, the model learns the correspondence between token sequences and valid molecular syntax, enabling it to generate chemically plausible molecules in a character-by-character manner \cite{bjerrum2017smiles, ozturk2020exploring}.

\subsection{Conditional Neural Processes}

Conditional Neural Processes (CNPs) provide a scalable alternative to Gaussian Processes (GPs) and Conditional Variational Autoencoders (CVAEs) by learning conditional distributions in a data-driven manner, avoiding restrictive priors \cite{Kim, Gordon}. They generalize well in low-data settings while maintaining computational efficiency with $O(n+m)$ complexity.

CNPs define a conditional stochastic process $Q_\theta$ over function values $f(x)$ given observations $O$:
\[
Q_\theta(f(T) \mid O, T) = \prod_{x \in T} Q_\theta(f(x) \mid O, x).
\]
The encoder $h_{\theta}$ maps observations $(x_i, y_i)$ into latent embeddings:
\begin{align*}
	r_i &= h_{\theta}(x_i, y_i), \quad r = \frac{1}{n} \sum_{i=1}^{n} r_i. 
\end{align*}
The decoder $g_{\theta}$ then predicts parameters $\phi_i$ for each target point $x_i \in T$:
\begin{align*}
	\phi_i &= g_{\theta}(x_i, r), \quad f(x_i) \sim \mathcal{N}(\mu_i, \sigma_i^2).
\end{align*}

CNPs are trained by minimizing the negative log-likelihood over randomly selected subsets of observations:
\begin{align*}
	\mathcal{L}(\theta) = -\mathbb{E}_{f \sim P} \left[ \mathbb{E}_{N} \log Q_{\theta} \left( \{y_i\}_{i=0}^{n-1} \mid O_N, \{x_i\}_{i=0}^{n-1} \right) \right]. 
\end{align*}
Monte Carlo methods and stochastic gradient descent (SGD) are used for optimization.

While CNPs offer efficient adaptation, they lack explicit uncertainty quantification, which can lead to overconfidence \cite{Jin W, Foong}. Effective subset selection strategies, such as adaptive sampling and curriculum learning, help mitigate variance and overfitting . CNPs are particularly well-suited for tasks requiring fast adaptation, such as online learning and molecular generation \cite{Radev}.

\subsection{Reptile Meta-Learning}

Reptile \cite{nichol2018first} is a simple yet effective first-order meta-learning algorithm designed for rapid adaptation to new tasks using limited data. Unlike Model-Agnostic Meta-Learning (MAML) \cite{finn2017model}, which optimizes an initialization for sensitivity to few gradient steps and often requires computing second-order gradients, Reptile works by repeatedly sampling a task, training on it for multiple steps using a standard optimizer, and then moving the initial model parameters towards the adapted parameters found for that task. This process implicitly finds a parameter initialization that is, on average, close to the optimal parameters for various tasks within the distribution, facilitating fast adaptation.

Given a task distribution \( \mathcal{P}(\mathcal{T}) \), each task \( \mathcal{T}_m \sim \mathcal{P}(\mathcal{T}) \) consists of a support set:
\[
\mathcal{D}_m^{\text{support}} = \{(x_i, y_i)\}_{i=1}^{N_s}
\]
and potentially a query set for evaluation (though not directly used in the Reptile update rule):
\[
\mathcal{D}_m^{\text{query}} = \{(x_j, y_j)\}_{j=1}^{N_q}.
\]

For each task \( \mathcal{T}_m \), the model parameters \( \theta \) are updated for \( k \) steps using the support set \( \mathcal{D}_m^{\text{support}} \) with a standard optimizer (like SGD or Adam) and an inner-loop learning rate \( \alpha \). Let \( U_{\mathcal{T}_m}^k(\theta) \) denote the parameters obtained after performing \( k \) optimization steps on task \( \mathcal{T}_m \) starting from \( \theta \). The task-adapted parameters are thus \( \theta_m' = U_{\mathcal{T}_m}^k(\theta) \).

The meta-update rule for Reptile involves moving the initial parameters \( \theta \) towards the task-adapted parameters \( \theta_m' \). Averaged over a batch of \( M \) tasks, the update is:
\begin{displaymath}
	\theta \leftarrow \theta + \epsilon \frac{1}{M} \sum_{m=1}^{M} (\theta_m' - \theta),
\end{displaymath}
where \( \epsilon \) is the outer-loop (meta) step size. This update can be interpreted as performing SGD on the objective \( \mathbb{E}_{\mathcal{T} \sim \mathcal{P}(\mathcal{T})} [ \frac{1}{2} || \theta - U_{\mathcal{T}}^k(\theta) ||^2 ] \).

Reptile's simplicity and effectiveness make it suitable for molecular generation, where adapting to new molecular property distributions or chemical spaces is crucial. Integrating Reptile into CNPs aims to enhance generalization with minimal fine-tuning, as outlined in Algorithm~\ref{algMetaMolGenReptile}.

\begin{algorithm}[h]
\caption{MetaMolGen Parameter Optimization via Reptile}
\label{algMetaMolGenReptile}
\begin{algorithmic}[1]
\Require Task distribution $\mathcal{P}(\mathcal{T})$, iteration number $T$, task batch size $M$
\Require Inner steps $k$, inner learning rate $\alpha$, outer step size $\epsilon$
\Require Initial parameters $\theta^{(0)}$
\Ensure Final meta-learned parameters $\theta^*$

\State Initialize $\theta \leftarrow \theta^{(0)}$
\For{$t = 1$ to $T$}
    \State Sample tasks $\{\mathcal{T}_m\}_{m=1}^M \sim \mathcal{P}(\mathcal{T})$
    \State Initialize $\Delta \theta \leftarrow 0$
    \For{each $\mathcal{T}_m$}
        \State Sample support set $\mathcal{D}_m^{\text{support}}$
        \State $\theta_m \leftarrow \theta$
        \For{$i = 1$ to $k$}
            \State $\mathcal{L}_m \leftarrow \mathcal{L}(\mathcal{D}_m^{\text{support}}; \theta_m)$
            \State $\theta_m \leftarrow \theta_m - \alpha \nabla_{\theta} \mathcal{L}_m$
        \EndFor
        \State $\Delta \theta \leftarrow \Delta \theta + (\theta_m - \theta)$
    \EndFor
    \State $\theta \leftarrow \theta + \epsilon \cdot \Delta \theta / M$
\EndFor
\State \Return $\theta$
\end{algorithmic}
\end{algorithm}

\section{Theoretical Analysis}
\label{SE5}

In this section, we provide a theoretical analysis of the proposed MetaMolGen model, focusing on its convergence properties and generalization guarantees. We establish conditions under which the training loss decreases monotonically and derive a PAC-Bayes bound to quantify generalization. 

All formal proofs of the following results are deferred to Appendix~\ref{app:proofs}.

\subsection{Assumptions}

We first introduce the fundamental assumptions that underlie our theoretical analysis.

\begin{assumption}[Lipschitz Smoothness \cite{nesterov2003introductory}]
\label{asm:smooth}
Let \( L_T(\theta) \) and the meta-learning loss \( \mathcal{L}(\theta) \) be differentiable functions with \( L \)-Lipschitz continuous gradients. Then, for all \( \theta_1, \theta_2 \in \mathbb{R}^d \),
\begin{align*}
\|\nabla L_T(\theta_1) - \nabla L_T(\theta_2)\| &\leq L \|\theta_1 - \theta_2\|, \\
\|\nabla \mathcal{L}(\theta_1) - \nabla \mathcal{L}(\theta_2)\| &\leq L \|\theta_1 - \theta_2\|.
\end{align*}
\end{assumption}

\begin{assumption}[Strong Convexity of Expected Loss]
\label{asm:strong-convex}
Let \( L_T(\theta) \) denote the task loss function and the expected loss \( \mathbb{E}_{T \sim p(T)} [L_T(\theta)] \) is \( \mu \)-strongly convex. Then, for all \( \theta \in \mathbb{R}^d \),
\[
L_T(\theta) \geq L_T(\theta^*) + \frac{\mu}{2} \|\theta - \theta^*\|^2,
\]
where \( \theta^* \) is the unique minimizer.
\end{assumption}

\subsection{Convergence Analysis}

To analyze the convergence behavior of MetaMolGen, we present three core theoretical results, each addressing a distinct aspect of training dynamics:

\begin{itemize}
\item \textbf{Theorem~\ref{TH1}} guarantees the stability and descent behavior of the optimization process under a properly chosen learning rate, contributing to improved training stability and time efficiency.

\item \textbf{Theorem~\ref{TH2}} demonstrates how input normalization reduces gradient variance, thereby improving the robustness and effectiveness of stochastic gradient updates. This also supports better diversity and faster convergence in molecular generation.

\item \textbf{Theorem~\ref{TH3}} shows that normalization significantly improves the conditioning of the loss landscape, which in turn accelerates convergence by reducing the dependency on the Hessian condition number. This has positive implications for time efficiency and enhances the model's ability to generate drug-like and synthesizable molecules.
\end{itemize}

These results elucidate how normalization techniques enhance training dynamics in MetaMolGen, contributing to both faster convergence and greater stability.

\begin{theorem}[Convergence of Training]
\label{TH1}
Let $L_T(\theta)$ be an $L$-smooth loss function. If the learning rate $\alpha$ satisfies
$0 < \alpha \leq \frac{2}{L},$
then the sequence of losses \( \{L_T(\theta_k)\} \) is monotonically decreasing, where

$$\theta_{k+1} = \theta_k - \alpha \nabla_{\theta} L_T(\theta_k).$$

\end{theorem}

\begin{theorem}[Gradient Variance Reduction via Normalization]
\label{TH2}
Let $X \in \mathbb{R}^d$ be the input molecular descriptors, $X' = \frac{X - \hat{\mu}}{\hat{\sigma} + \epsilon}$ denote their normalized version. $\theta$ and $\theta'$ are model parameters obtained by training on $X$ and $X'$ , respectively. Then,
$$\operatorname{Var}(\nabla_\theta L_T(\theta')) \leq \operatorname{Var}(\nabla_\theta L_T(\theta)).$$

\end{theorem}

\begin{theorem}[Improved Conditioning and Accelerated Convergence]
\label{TH3}
Let $L_T(\theta)$ be twice differentiable, with Hessian $H = \nabla^2_\theta L_T(\theta)$ and condition number $\kappa(H) = \lambda_{\max}(H) / \lambda_{\min}(H)$. Consider normalized inputs $$X' = (X - \hat{\mu}) / (\hat{\sigma} + \epsilon),$$ and let $L_T'(\theta)$ be the corresponding loss with Hessian $H' = \nabla^2_\theta L_T'(\theta)$. Then, $$\kappa(H') \ll \kappa(H),$$ and the iteration complexity of first-order optimization improves from $O\left(\kappa(H) \log \tfrac{1}{\epsilon}\right)$ to $O\left(\log \tfrac{1}{\epsilon}\right)$.
\end{theorem}

\subsection{Generalization and Error Analysis}

To analyze the generalization behavior of MetaMolGen, we present four theoretical results, each isolating a key contributor to the total expected loss and characterizing a distinct source of error:

\begin{itemize}
\item \textbf{Theorem~\ref{TH4}} shows that input normalization reduces meta-generalization error, highlighting its role in enhancing generalization under distributional shifts. This improves validity, uniqueness, and diversity in generated molecules.

\item \textbf{Theorem~\ref{TH5}} establishes the unbiasedness of the stochastic gradient estimator used in MetaMolGen, ensuring the reliability of updates computed from small mini-batches, which is essential for efficient training under data-scarce conditions.

\item \textbf{Theorem~\ref{TH6}} provides a convergence guarantee under stochastic optimization, bounding the expected suboptimality and quantifying the impact of variance. This supports more stable training and indirectly benefits the quality of drug-like and synthesizable outputs.

\item \textbf{Theorem~\ref{TH7}} offers a unified error bound that decomposes the total expected loss into four components: optimization error, generalization error, gradient noise, and model approximation error. 
\end{itemize}

The results characterize the error dynamics of MetaMolGen and clarify how training stability, generalization, and model capacity interact in few-shot molecular generation.

\begin{theorem}[Variance Reduction and Generalization Improvement]
\label{TH4}
Let $X \in \mathbb{R}^d$ be molecular descriptors, and $\mathcal{E}_{\text{meta, raw}}$, $\mathcal{E}_{\text{meta, standardized}}$ be the meta-generalization errors under training on $X$ and its normalized form $X' = (X - \hat{\mu}) / (\hat{\sigma} + \epsilon)$, respectively. Then,
$$
\mathbb{E}[\mathcal{E}_{\text{meta, standardized}}] < \mathbb{E}[\mathcal{E}_{\text{meta, raw}}].
$$
\end{theorem}

\begin{theorem}[Unbiasedness of Stochastic Gradient]
\label{TH5}
Let $\mathcal{D}_T = \{(x_i, z_i, y_i)\}_{i=1}^{N_T}$ be the dataset for task $T$, where $x_i$ is the molecular input, $z_i$ is the task-specific context, and $y_i$ is the corresponding label. Let $\mathcal{B} \subset \mathcal{D}_T$ be a mini-batch sampled uniformly. The stochastic gradient estimator of $L_T(\theta)$ is defined as
$$
\nabla_{\text{est}} L_T(\theta) = \frac{1}{|\mathcal{B}|} \sum_{(x_i, z_i, y_i) \in \mathcal{B}} \nabla_\theta \mathcal{L}(f_\theta(x_i, z_i), y_i),
$$
where $f_\theta$ is the model and $\mathcal{L}$ is the sample-wise loss. Then the estimator is unbiased:
$$
\mathbb{E}_{\mathcal{B}}[\nabla_{\text{est}} L_T(\theta)] = \nabla_\theta L_T(\theta).
$$
\end{theorem}

\begin{theorem}[Convergence under Stochastic Updates]
\label{TH6}
Let $L_T(\theta)$ be an $L$-smooth loss function, and the stochastic gradient estimator is unbiased with bounded variance $\sigma^2$. The parameter after $k$ iterations of SGD is denoted by $\theta_k$, and $\theta^*$ be the minimizer of $L_T(\theta)$. Then the expected suboptimality satisfies:
$$
\mathbb{E}[L_T(\theta_k)] - L_T(\theta^*) = O\left( \frac{L}{k} \right).
$$
\end{theorem}

\begin{theorem}[Generalization and Error Bound]
\label{TH7}
Let \(\theta_k\) be the model parameters after \(k\) updates, and the optimal parameters minimizing expected loss is denoted by $\theta^*$. Then the total expected error satisfies:
\begin{align*}
\mathbb{E}_{T \sim p(T)}[L_T(\theta_k)] - L_T(\theta^*) 
&\leq O\left( \frac{L}{k} \right) + O\left( \sqrt{ \frac{D_{\text{KL}}(Q \| P)}{N} } \right) \\
&\quad + O\left( \frac{\sigma}{\sqrt{B}} \right) + O(\epsilon_{\text{approx}}),
\end{align*}
where the four terms represent: 
$O\left( \frac{L}{k} \right)$ for optimization error, 
$O\left( \sqrt{ \frac{D_{\text{KL}}(Q \| P)}{N} } \right)$ for generalization error, 
$O\left( \frac{\sigma}{\sqrt{B}} \right)$ for gradient noise, 
and $O(\epsilon_{\text{approx}})$ for approximation error.

\end{theorem}

\section{Experimental Setup} \label{SE6}

\subsection{Datasets}  
MetaMolGen is evaluated on multiple benchmark datasets covering diverse molecular structures and properties. The model is trained on ChEMBL \cite{Gaulton2012chembl} and tested on QM9 \cite{Ramakrishnan2014}, ZINC \cite{Irwin2012}, and MOSES \cite{Polykovskiy2020}. These datasets provide a comprehensive platform for assessing molecular generation in low-data scenarios, ensuring that the model is capable of producing valid, novel, and structurally diverse molecules while optimizing key physicochemical properties. A subset of $1,000$ to $60,000$ molecules is selected from each dataset to simulate a few-shot learning setting, where data availability is inherently constrained.  

\subsection{Baselines}  
MetaMolGen is evaluated against several advanced molecular generative models, including MolGAN , an implicit graph generative model based on generative adversarial networks (GANs) and reinforcement learning (RL), recurrent neural network-based models (RNNs), and transformer-based architectures. These approaches represent mainstream paradigms in molecular design, yet they often struggle to capture distributional patterns effectively in low-data regimes. The experimental comparisons underscore the need for a more efficient and adaptable generative framework.
 
\subsection{Model Configuration}  
MetaMolGen performs molecular sequence generation under few-shot settings, as outlined in Algorithm~\ref{alg:MetaMolGenForward}. Input molecular features are first standardized, then encoded into latent representations by the context encoder. These are mean-pooled and refined by the task encoder to produce a global embedding, which conditions a two-layer LSTM decoder to generate SMILES sequences.

\subsubsection{Feature Normalization.}
The feature normalization layer standardizes molecular descriptors to zero mean and unit variance within each batch, improving numerical stability and training consistency. This normalization is applied both during forward generation and inner-loop adaptation:
$X' = \frac{X - \hat{\mu}}{\hat{\sigma} + \epsilon}.$
Here, $\hat{\mu}$ and $\hat{\sigma}$ are computed across the batch. Beyond improving optimization, this transformation also contributes to generalization by reducing gradient noise and input variability, particularly helpful in low-data molecular tasks (see Theorem~\ref{TH4}).

\subsubsection{Architecture Design.}
The context encoder is a three-layer fully connected network with GELU activations, reducing feature dimensionality to extract molecular representations. The hidden and latent dimensions are set to 256 and 128, respectively. Mean pooling is used to aggregate context representations, producing a global task-level embedding that is refined by the task encoder through linear transformation and layer normalization.

The SMILES decoder is a two-layer LSTM (hidden/embedding dimension: 128), which autoregressively generates molecular sequences from the task representation. Layer normalization stabilizes the decoder’s hidden states, and dropout (0.2) is applied to mitigate overfitting.

\subsubsection{Optimization Settings.}
MetaMolGen is trained using the Reptile meta-learning algorithm. Each training iteration samples a small batch of molecular subsets to construct tasks. Task-specific parameters are adapted via inner-loop gradient updates (learning rate: 0.01), and the shared initialization is then updated using Reptile's first-order meta-update rule. Meta-optimization is performed using the Adam optimizer (initial learning rate: 0.001, weight decay: 0.01). Training is conducted for 150 epochs, with 10 tasks per meta-update and 16 molecules per task. Dropout and $L_2$ regularization are applied throughout to improve generalization.

\subsubsection{Batch Size and Gradient Variance.}
To reduce gradient estimation noise, we use a mini-batch size of 64. This setting balances variance reduction (captured theoretically as $O\left(\frac{\sigma}{\sqrt{B}}\right)$ in Theorem~\ref{TH7}) and computational efficiency. Gradient noise in few-shot settings comprises within-task variance and across-task gradient drift. Increasing batch size suppresses the former, while averaging over tasks mitigates inter-task fluctuations, jointly enhancing update stability.

\subsection{Evaluation Measures}  
To summarize performance across multiple objectives, we define an \textit{Overall Score} as the unweighted average of normalized values from the following seven metrics: 
(1) Validity, 
(2) Uniqueness,
(3) Time efficiency, 
(4) Diversity, 
(5) Druglikeness, 
(6) Synthesizability
(7) Solubility.

Each metric is normalized based on its observed minimum and maximum values across all baseline models and the evaluated model. For metrics where higher values indicate better performance (i.e., Validity, Uniqueness, Diversity, Druglikeness, Synthesizability, and Solubility), we apply min-max normalization as:
\[
\text{Normalized}_i = \frac{x_i - x_i^{\min}}{x_i^{\max} - x_i^{\min}}.
\]
For metrics where lower values are preferred (i.e., Time), the normalization is inverted:
\[
\text{Normalized}_i = \frac{x_i^{\max} - x_i}{x_i^{\max} - x_i^{\min}}.
\]
If a metric exhibits zero range across models (i.e., \(x_i^{\max} = x_i^{\min}\)), the normalized score is set to 0.5 to avoid division by zero.

The final Overall Score is computed as the mean of the normalized metrics:
\begin{equation}
\text{Overall Score} = \frac{1}{7} \sum_{i=1}^{7} \text{Normalized}_i
\label{eq:overall_score}
\end{equation}

\begin{algorithm}[htbp]
\caption{MetaMolGen Forward Molecular Generation Process}
\label{alg:MetaMolGenForward}
\begin{algorithmic}[1]
\Require Context features $X \in \mathbb{R}^{B \times N \times d}$, input SMILES $S_{\text{in}} \in \mathbb{N}^{B \times L}$, model parameters $\theta$
\Ensure Predicted token logits $P \in \mathbb{R}^{B \times L \times V}$
\State Normalize features: $\hat{X} \leftarrow \dfrac{X - \mu}{\sigma + \epsilon}$
\State Encode context points: $H \leftarrow \text{ContextEncoder}(\hat{X})$
\State Aggregate context: $r \leftarrow \text{mean}(H, \text{dim}=1)$
\State Refine task representation: $r_{\text{task}} \leftarrow \text{TaskEncoder}(r)$
\State Generate SMILES logits: $P \leftarrow \text{SMILESDecoder}(S_{\text{in}}, r_{\text{task}})$
\Return $P$
\end{algorithmic}
\end{algorithm}

\subsection{Ablation Study}  
To assess the impact of the standardization layer, we compare MetaMolGen’s performance on molecular generation across dataset sizes (1,000 to 60,000 molecules) with and without feature normalization. Results show that removing the standardization layer reduces validity, novelty, and diversity while increasing errors in property optimization, especially in low-data settings. This highlights its role in stabilizing feature distributions, improving generalization, and enhancing robustness in few-shot molecular generation.

For details on datasets, baselines, and hyperparameter configurations, please refer to \url{https:}.

\section{Empirical Results}\label{SE7}

We conduct a comprehensive evaluation of MetaMolGen across multiple dimensions, including comparison with competitive baseline models, performance under few-shot learning scenarios, an ablation study of the feature standardization component, as well as assessments of property control and structural diversity. The experiments are performed on both standard molecular generation benchmarks and customized few-shot configurations, covering a wide range of data availability settings.

MetaMolGen consistently outperforms strong baselines in terms of generation quality, structural diversity, and generalization under limited data conditions. The incorporation of the feature standardization layer proves critical for enhancing the model’s training stability and improving alignment with target properties. Furthermore, MetaMolGen demonstrates robust conditional generation capabilities, effectively steering molecular outputs toward desired property values (e.g., \textit{logP}, \textit{TPSA}, \textit{SAS}, and \textit{QED}) while preserving molecular diversity.

Overall, these results highlight the model’s ability to balance validity, controllability, and diversity across varying data regimes, establishing MetaMolGen as a promising framework for data-efficient and property-aware molecular design.

\subsection{Comparison with Baseline Models}

To comprehensively evaluate the performance of MetaMolGen, we compare it with mainstream baseline models, including ORGAN~\cite{Guimaraes2017}, MolGAN~\cite{De Cao}, RNN~\cite{Segler}, and MolGPT~\cite{Bagal2022}. Following the original experimental settings, we use a dataset of 5,000 molecules for both training and generation across all models. 

\begin{figure}[htbp]
    \centering
    \includegraphics[width=0.48\textwidth]{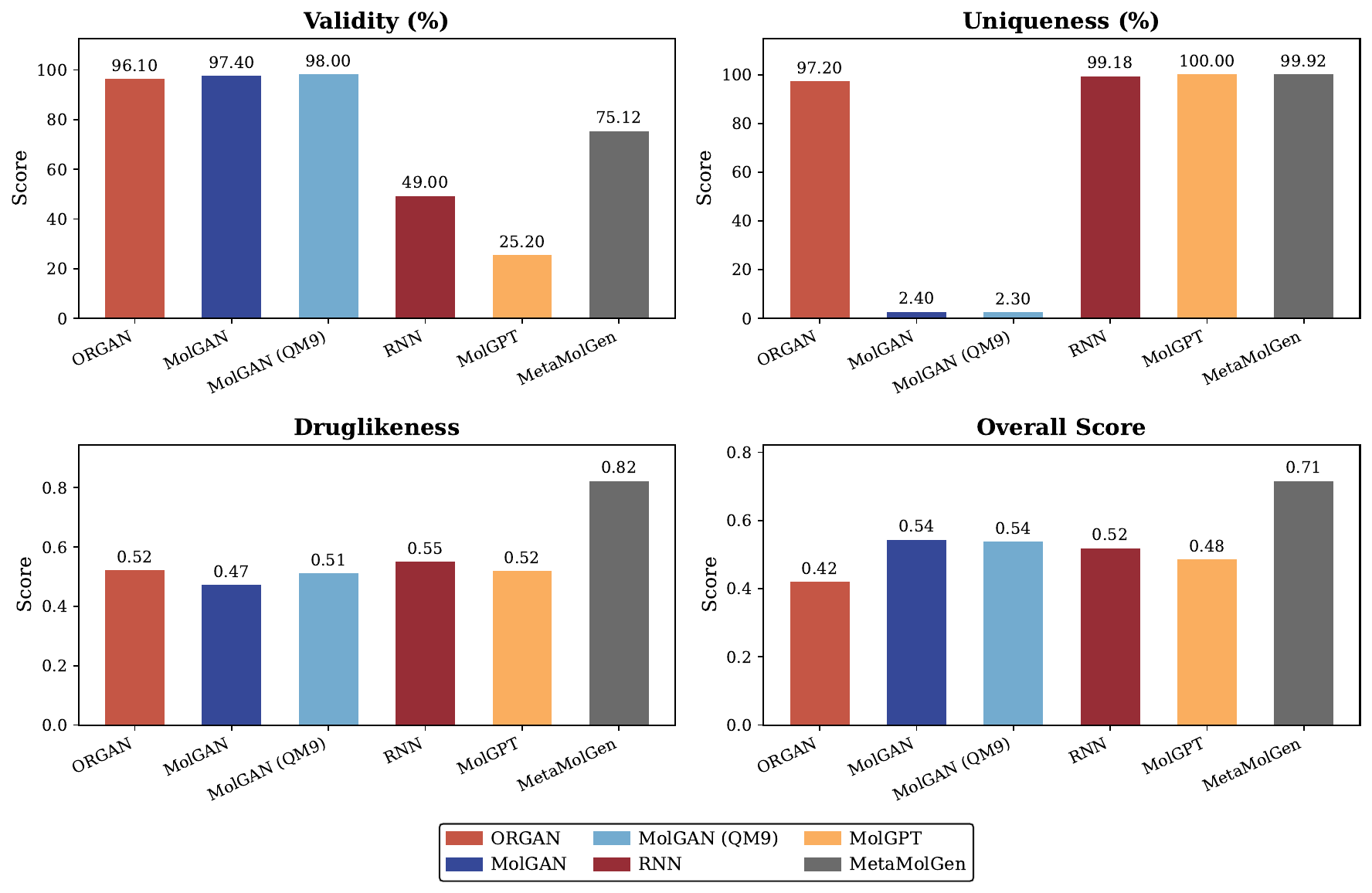}
    \caption{Comparison of key generation metrics across different models, including validity, uniqueness, druglikeness, and overall performance score.}
    \label{fig:model_metrics_barplot}
\end{figure}

As shown in Table~\ref{tab:model_comparison} and Figure~\ref{fig:model_metrics_barplot}, MetaMolGen achieves superior performance across key molecular generation metrics. Although its validity is slightly lower than reinforcement learning-based models such as ORGAN and MolGAN, it substantially outperforms all baselines in uniqueness and essential drug-related properties.

To summarize overall performance, we define an Overall
 Score as a weighted average of normalized values across several key metrics, including validity, uniqueness, diversity, and druglikeness.  MetaMolGen achieves the highest overall score, highlighting its balanced optimization of chemical quality, diversity, and efficiency.

These results demonstrate that MetaMolGen strikes a better balance between molecular property optimization and structural diversity, while maintaining excellent generation efficiency. In contrast, reinforcement learning models prioritize validity but often fail to ensure uniqueness and chemical diversity. Sequence models, although computationally efficient, lack robustness in producing chemically valid structures under the same experimental settings.

\begin{table*}[h]
\centering
\caption{Comparison of molecular generation performance across models. Gray cells indicate metrics where our model outperforms baselines.}
\small
\resizebox{\textwidth}{!}{
\begin{tabular}{llcccccccc}
\hline
Objective & Algorithm & Valid (\%) & Unique (\%) & Time (h) & Diversity & Druglikeness & Synthesizability & Solubility & Overall Score \\
\hline
All/Alternated & ORGAN & 96.1 & 97.2* & 10.2* & 0.92 & 0.52 & 0.71 & 0.53 & 0.4183 \\
All/Simultaneously & MolGAN & 97.4 & 2.4 & 2.12 & 0.91 & 0.47 & 0.84 & 0.65 & 0.5419 \\
All/Simultaneously & MolGAN (QM9) & \cellcolor{gray!25}98.0 & 2.3 & 5.83 & \cellcolor{gray!25}0.93 & 0.51 & 0.82 & 0.69 & 0.5356 \\
Sequence-based & RNN & 49.0 & 99.18 & 1.50 & 0.8742 & 0.5487 & 0.8104 & 0.6849 & 0.5170 \\
Sequence-based & MolGPT & 25.2 & \cellcolor{gray!25}100.0 & 1.50 & 0.8637 & 0.5162 & 0.8213 & 0.6440 & 0.4836 \\
All/Simultaneously & \cellcolor{gray!25}MetaMolGen & 75.12 & 99.92 & \cellcolor{gray!25}0.05 & 0.8337 & \cellcolor{gray!25}0.8206 & \cellcolor{gray!25}0.8553 & \cellcolor{gray!25}0.8876 & \cellcolor{gray!25}0.7143 \\
\hline
\end{tabular}
}
\label{tab:model_comparison}
\end{table*}

\vspace{0.5em}
\noindent

\subsection{Generation Performance on Small-Sized Datasets}

As our preliminary experiments revealed that MolGAN performs poorly in terms of uniqueness, we excluded it from the few-shot evaluation. Instead, we selected three models with relatively good uniqueness performance—MetaMolGen, RNN, and MolGPT—for systematic evaluation under low-resource conditions, using training set sizes ranging from 1,000 to 10,000 samples.

\begin{figure*}[h]
\centering
\includegraphics[width=0.8\textwidth]{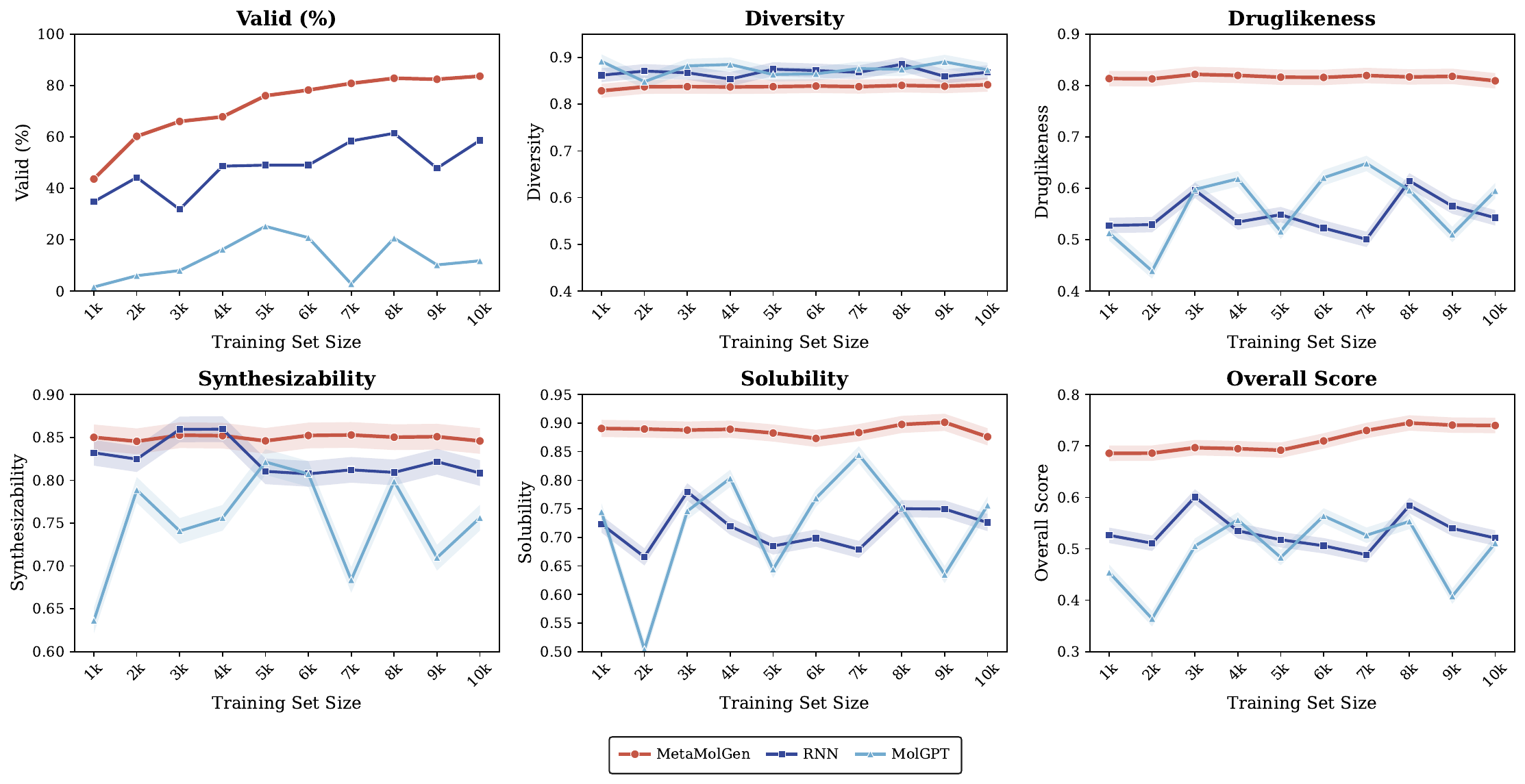}
\caption{Few-shot performance comparison of MetaMolGen, RNN, and MolGPT across training set sizes (1k–10k) on six key metrics: validity, diversity, drug-likeness, synthesizability, solubility, and overall score.}
\label{fig:fewshot_metrics}
\end{figure*}

As shown in the key performance trends illustrated in Figure~\ref{fig:fewshot_metrics}, MetaMolGen consistently outperforms both RNN and MolGPT across all training sizes. The advantage is particularly evident in extremely low-data scenarios, where MetaMolGen demonstrates strong stability and validity. Furthermore, MetaMolGen maintains excellent and consistent performance across other important properties, including drug-likeness, synthesizability, solubility, and overall generation quality.

In contrast, RNN exhibits notable fluctuations across training sizes, while MolGPT, although competitive in diversity and uniqueness, suffers from significantly low validity and limited overall generation quality.

A detailed breakdown of metric scores across all training sizes is provided in Appendix C, Table~\ref{tab:fewshot_valid}.

\subsection{Conditional Property Control}

To evaluate the controllability and diversity of molecule generation, we tested the MetaMolGen model under conditional generation settings using four widely known compounds as property targets: Aspirin, Tamiflu, Amoxicillin, and Chloroquine. These compounds represent therapeutic classes spanning anti-inflammatory, antiviral, antibacterial ($\beta$-lactam antibiotics), and antiparasitic domains, thus offering a broad testbed for conditional molecule generation.

The target properties include molecular weight, logarithm of the partition coefficient, the number of hydrogen bond donors, and hydrogen bond acceptors. To encourage structural diversity while maintaining property alignment, Gaussian noise was injected into the latent space during generation. The model was trained on a dataset of 5,000 molecules, following the same setup as MolGAN.

To enable explicit control over molecular properties, we introduced a lightweight MLP-based property conditioning module into the MetaMolGen architecture. This module projects the target property vector into the latent space and injects it into the initial hidden state of the LSTM decoder. This design allows property-guided sequence generation in an end-to-end trainable manner.

\begin{figure*}[h]
	\centering
	\includegraphics[width=0.9\textwidth]{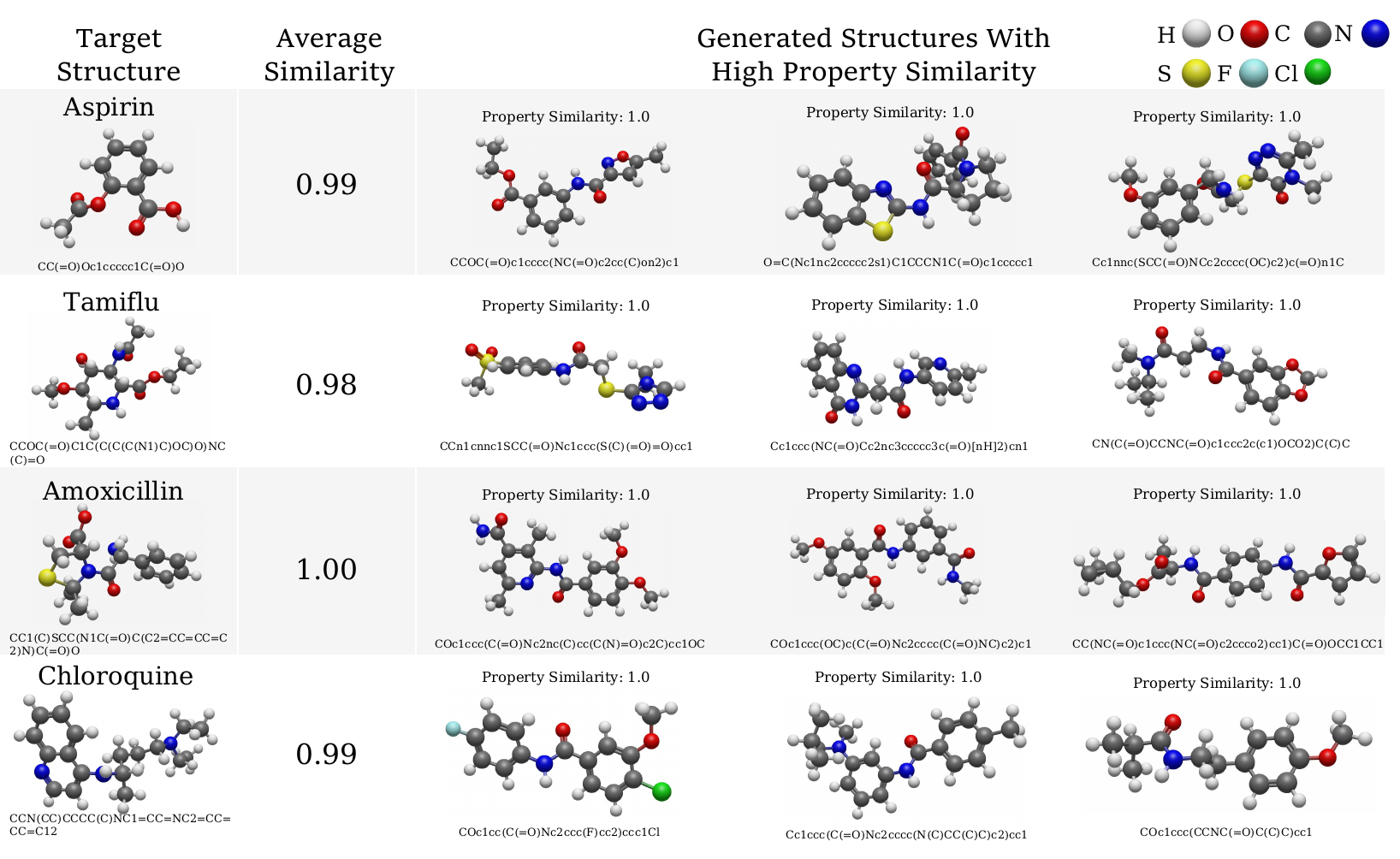} 
	\caption{Representative molecules generated by the MetaMolGen model using the condition vectors of Aspirin, Tamiflu, Amoxicillin, Chloroquine. Molecules exhibit high property alignment.}
	\label{fig:1a}
\end{figure*}

Figure~\ref{fig:1a} shows representative molecules generated under the property conditions of the four compounds. Notably, all displayed molecules achieve a property similarity of 1.0, indicating perfect alignment across the specified property dimensions. The average property similarities achieved were 0.99 for Aspirin, 0.98 for Tamiflu, 1.00 for Amoxicillin, and 0.99 for Chloroquine, demonstrating the model’s precise conditional generation capability.

Further statistics show that the mean absolute error (MAE) in logP was 0.15, while the MAE in molecular weight (MW) was 0.32, both of which are closely aligned with the target values. Compared to baseline models lacking explicit conditioning (e.g., MolGPT), MetaMolGen exhibits superior property alignment and structural fidelity. The model not only preserves core chemical characteristics but also produces valid and diverse structures.

\begin{figure}[htbp]
\centering
\includegraphics[width=\linewidth]{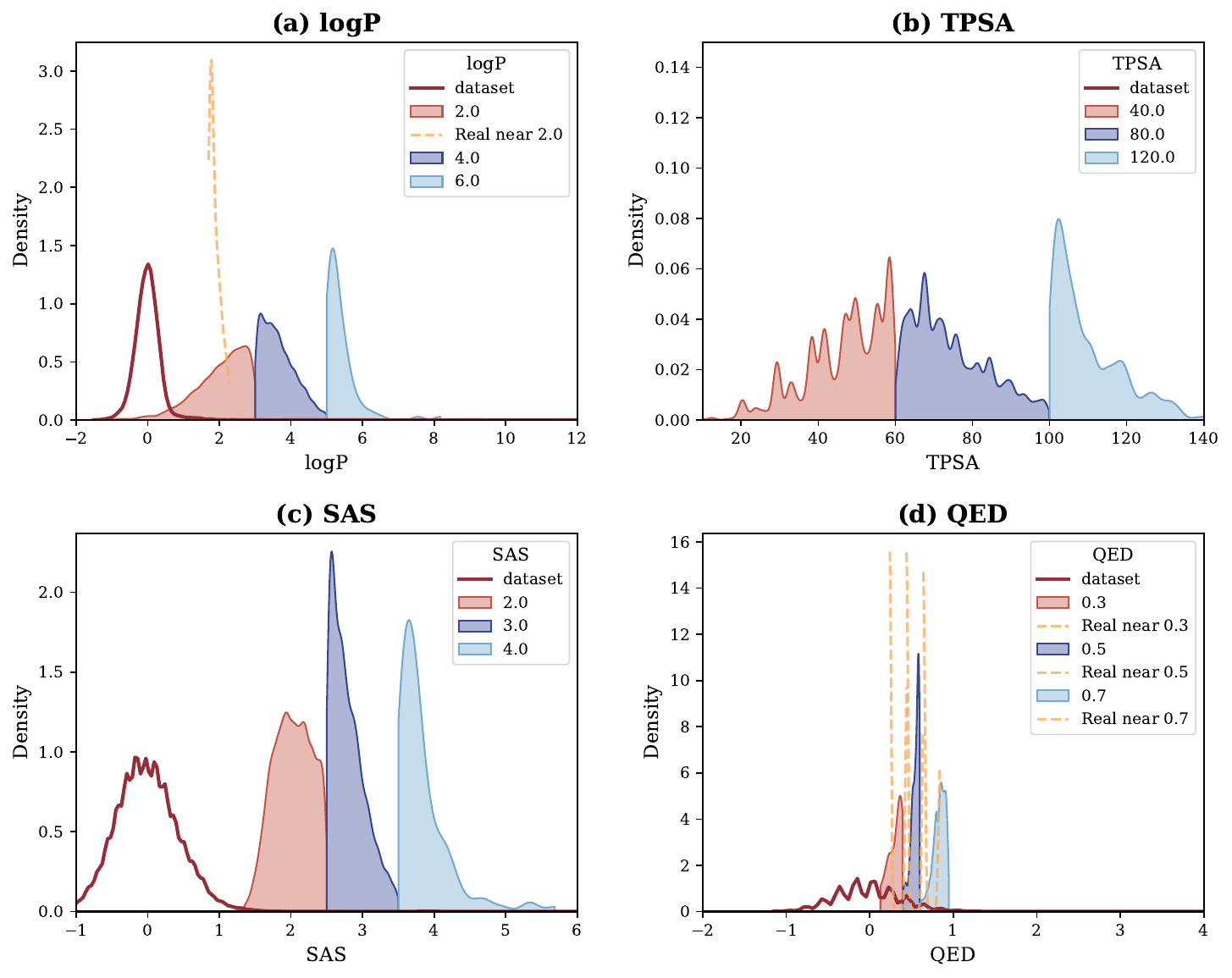}
\caption{
Distributions of generated molecular properties under different target constraints: (a) logP, (b) TPSA, (c) SAS, and (d) QED. 
Shaded regions indicate the generated samples under specific property conditions. 
Solid red lines show the distribution of the full real dataset, while dashed lines correspond to real molecules near the target values.
The close alignment between generated and real-near-target distributions demonstrates the effectiveness of the proposed conditional control mechanism.
}
\label{fig:property_distributions}
\end{figure}

As shown in Figure~\ref{fig:property_distributions}, MetaMolGen demonstrates robust control across multiple property dimensions. The property distributions of generated molecules form unimodal peaks closely aligned with those of real molecules near the same target values. Particularly strong control is observed for QED and SAS. Although TPSA exhibits higher variance and is more challenging to constrain, the overall trend remains aligned with the targets.

Compared to baseline models without explicit property conditioning mechanisms (such as MolGPT), MetaMolGen achieves enhanced controllability and improved structural alignment under property constraints.

\subsection{Ablation Study: Effect of Standardization Layer}

\begin{figure}[htbp]
    \centering

    \begin{subfigure}[b]{0.9\linewidth}
        \centering
        \includegraphics[width=\linewidth]{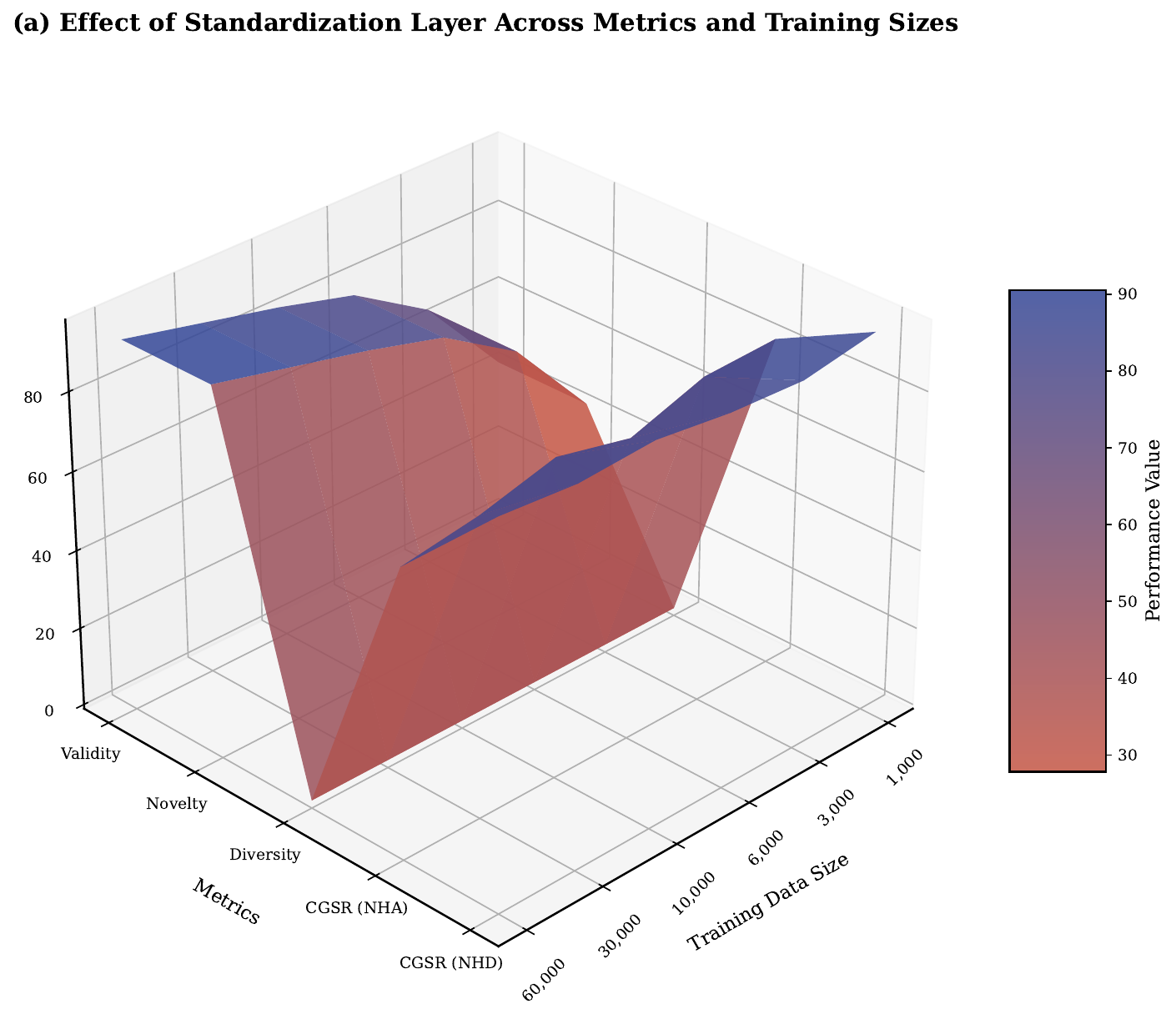}
        \caption{Effect of the standardization layer across multiple evaluation metrics under varying training data sizes.}
        \label{fig:ablation_surface}
    \end{subfigure}
    
    \vspace{0.8em}  

    \begin{subfigure}[b]{0.9\linewidth}
        \centering
        \includegraphics[width=\linewidth]{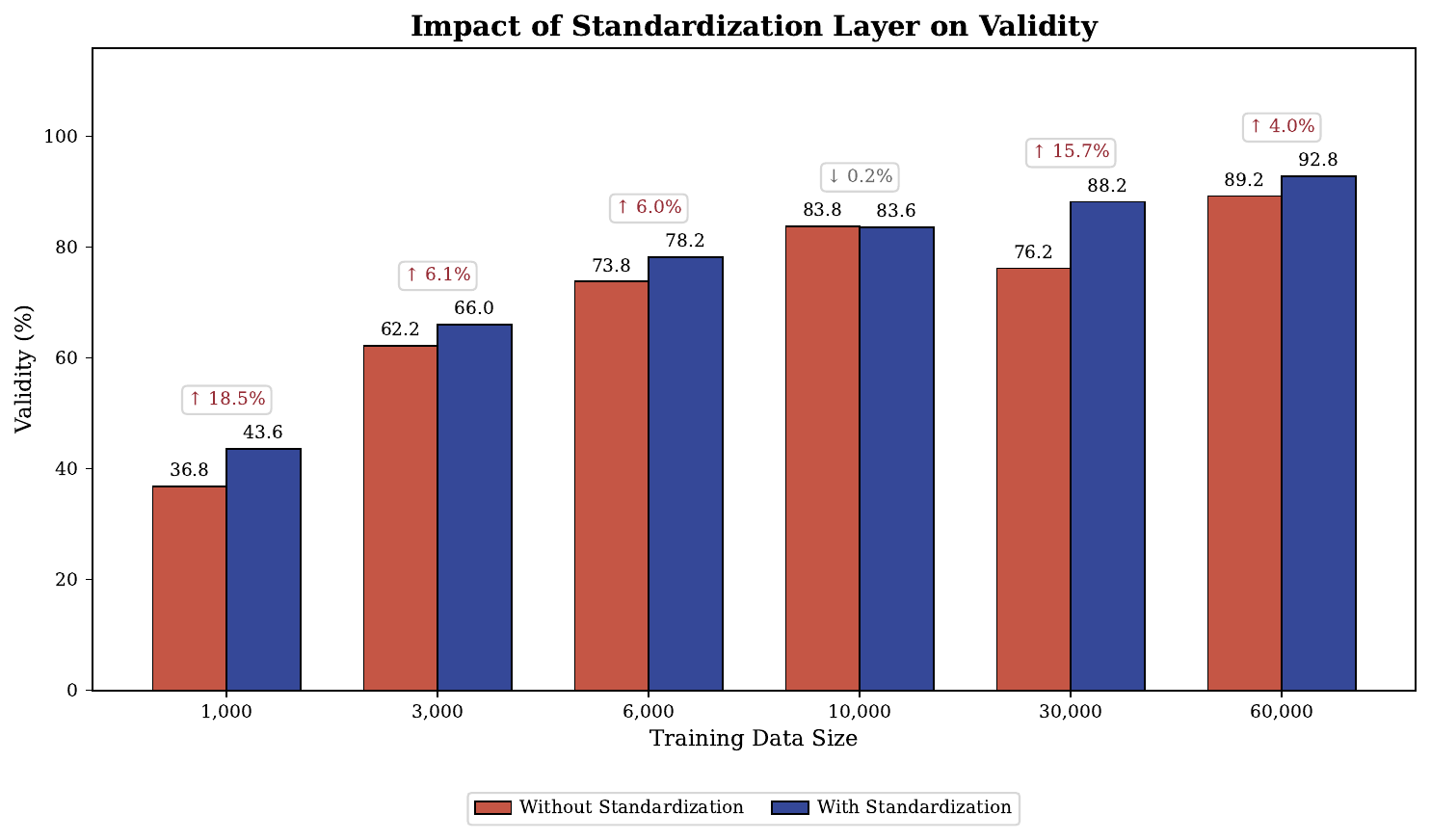}
        \caption{Impact of the standardization layer on validity under different training data sizes. Blue bars: with standardization; red bars: without. Improvements are especially evident under low-data regimes.}
        \label{fig:ablation_bar}
    \end{subfigure}

    \caption{Ablation study on the effect of the standardization layer across training sizes.}
    \label{fig:ablation_combined}
\end{figure}

We conducted an ablation study to evaluate the role of the feature standardization module in our model. As shown in the visual comparisons in Figure~\ref{fig:ablation_combined}, removing the standardization layer leads to noticeable declines in key performance metrics, including validity, novelty, and conditional generation success rate (CGSR). These degradations are particularly pronounced under low-data conditions, highlighting that feature standardization plays a crucial role in maintaining generation quality.

A detailed numerical comparison across training sizes and metrics is provided in Appendix C, Table~\ref{tab:ablation_study}.

Overall, the feature standardization module not only helps improve the chemical correctness and property alignment of generated molecules, but also accelerates model convergence and reduces training instability. These results underscore the importance of appropriate preprocessing strategies in enhancing robustness and generalization under few-shot settings.

\section{Model Discussion}\label{SE8}
Through the integration of feature standardization, conditional generation mechanisms, and meta-learning strategies, MetaMolGen achieves a strong balance of generation quality, stability, and data efficiency. Its robust performance across multiple property constraints and its ability to generate diverse, drug-like molecules make it a promising framework for data-efficient, goal-driven molecular design.
\subsection{Model Generation and Conditional Generation Capabilities}

MetaMolGen's molecular generation ability arises from its context-aware modeling and the integration of attribute conditioning. The core components, including the Context Encoder and Aggregator, allow the model to learn global task representations from molecular features. The Property Projector maps target properties to a latent space, which is then used to guide the decoding process, ensuring that the generated molecules adhere to the desired property constraints.

From the overall performance comparison in Table~\ref{tab:model_comparison}, MetaMolGen demonstrates superior performance in several key metrics. While its validity (75.12\%) is slightly lower than that of MolGAN (97.4\%), it outperforms all other models in several critical aspects, including \textit{Druglikeness} (0.8206), \textit{Synthesizability} (0.8553), and \textit{Solubility} (0.8876), making the generated molecules more suitable for real-world applications in drug design. Moreover, the model's generation time is only 0.05 hours, a significant improvement over ORGAN (10.2 hours) and MolGAN (2.12 hours), highlighting MetaMolGen’s efficiency in molecular generation tasks.

In conditional generation, MetaMolGen demonstrates the ability to effectively condition the generation process on various molecular properties. As shown in Table~\ref{tab:split_condition_model}, the model consistently maintains high novelty and uniqueness across multiple property constraints, such as LogP, TPSA, SAS, and QED, with values above 94.9\%. However, the mean absolute deviation (MAD) and standard deviation (SD) are slightly higher compared to MolGPT, indicating that the model still faces challenges in perfectly aligning the generated molecules with the target properties, but it nonetheless generates molecules with more diverse and realistic chemical properties.

\begin{table}[htbp]
\footnotesize 
\caption{Performance comparison under different property constraints (metrics adapted from Table 4 in~\cite{Bagal2022}).}
\label{tab:split_condition_model}
\begin{tabularx}{\linewidth}{l l *{5}{>{\centering\arraybackslash}X}}
\toprule
\textbf{Condition} & \textbf{Model} & \textbf{Val.} & \textbf{Uni.} & \textbf{Nov.} & \textbf{MAD} & \textbf{SD} \\
\midrule
LogP   & MolGPT      & 0.971 & 0.998 & 1.000 & 0.23   & 0.31   \\
       & MetaMolGen  & 0.949 & 0.993 & 0.997 & 1.54   & 0.89   \\
\addlinespace
TPSA   & MolGPT      & 0.972 & 0.996 & 1.000 & 3.52   & 4.66   \\
       & MetaMolGen  & 0.949 & 0.993 & 0.997 & 123.40 & 31.20  \\
\addlinespace
SAS    & MolGPT      & 0.977 & 0.995 & 1.000 & 0.13   & 0.20   \\
       & MetaMolGen  & 0.949 & 0.993 & 0.997 & 0.57   & 0.61   \\
\addlinespace
QED    & MolGPT      & 0.975 & 0.998 & 1.000 & 0.056  & 0.075  \\
       & MetaMolGen  & 0.949 & 0.993 & 0.997 & 1.00   & 0.93   \\
\bottomrule
\end{tabularx}
\end{table}
\subsection{Impact of Feature Standardization on Model Robustness}

Feature standardization plays a crucial role in improving the robustness and stability of MetaMolGen. By normalizing the molecular features through a learnable mean and standard deviation, the model is able to mitigate the impact of feature scale differences, ensuring that the model performs consistently across various tasks, data sizes, and molecular representations.

As shown in Table~\ref{tab:model_comparison}, MetaMolGen outperforms models such as RNN (Validity: 49.0\%) and Transformer (Validity: 25.2\%) in terms of both validity (75.12\%) and overall score (0.7143). This improvement can be attributed to the inclusion of the feature normalization layer, which stabilizes training, improves gradient flow, and enhances property alignment during the optimization process. Furthermore, feature standardization contributes to better convergence during training, particularly in low-data regimes where other models struggle to achieve stable results.

The incorporation of standardization not only improves the model's overall performance but also addresses issues such as gradient explosion and feature dominance, particularly in tasks that require attribute control. This stabilization mechanism provides a strong foundation for MetaMolGen's ability to learn complex tasks with minimal supervision, making it a highly adaptable model for a wide range of molecular design applications.

\subsection{Model Efficiency and Data Utilization}

MetaMolGen’s architecture, which combines feature standardization with a task-level modeling mechanism, excels in data efficiency, especially when data is scarce. As seen in Table~\ref{tab:model_comparison}, the model achieves impressive performance with a generation time of only 0.05 hours, far outperforming other models such as ORGAN (10.2 hours) and MolGAN (2.12 hours). This computational efficiency is critical in real-world applications where time and resources are limited.

Moreover, MetaMolGen leverages meta-learning principles to enable fast adaptation to new tasks with minimal data. The Task Encoder and Aggregator modules enable the model to quickly learn and adapt to novel tasks by aggregating information from limited data points, further enhancing its generalization ability. Despite slightly higher MAD in certain conditional generation tasks (e.g., TPSA and QED), MetaMolGen maintains high success rates and uniqueness (Validity: $>$ 94.9\%, Uniqueness: $>$ 99.3\%), demonstrating its robust data utilization capability.

Overall, MetaMolGen proves to be a highly efficient and versatile model, capable of quickly adapting to new molecular generation tasks with limited data while maintaining high performance in terms of both generation quality and property control.

\section{Conclusions and Limitations}
\label{SE9}

\subsection{Technical Discussions}

This work proposes MetaMolGen, a meta-learning-based framework designed to address few-shot molecular generation and multi-objective optimization under data scarcity. Leveraging the first-order meta-learning algorithm Reptile, MetaMolGen efficiently adapts to new molecular tasks with limited supervision, improving generalization while maintaining both structural validity and alignment with target properties. 

Experimental evaluations on ChEMBL, QM9, ZINC, and MOSES demonstrate the model's strong performance in low-data regimes. With only 1,000–10,000 training samples, MetaMolGen consistently achieves high validity, novelty, and diversity, and exceeds 95\% conditional generation success rate (CGSR) under hydrogen bond donor/acceptor (HBD/HBA) constraints, outperforming conventional baselines.

Beyond generative performance, MetaMolGen also enhances molecular screening efficiency by reducing computational overhead through its controllable generation mechanism. This is enabled by a learnable property projector that integrates target attributes (e.g., LogP, TPSA, QED, SAS) into the latent space, allowing the model to generate property-aligned molecules directly, without reliance on post-hoc filtering.

\subsection{Existing Limitations}

While MetaMolGen demonstrates strong adaptability and generalization, several limitations remain. First, task-specific adaptation in latent space is still coarse-grained, potentially limiting fine-grained property tuning in complex design scenarios. Second, the chemical viability of some generated molecules has not yet been fully validated under real-world synthesis constraints. Finally, while conditional control via latent projection is effective, it may require retraining when generalizing to entirely new property domains.

\subsection{Future Extensions}

Future work will focus on several key directions. First, we aim to explore advanced meta-learning algorithms for fine-grained control over molecular attributes, including gradient-based and metric-based adaptation hybrids. Second, we plan to integrate active learning strategies to further reduce data requirements and improve sample efficiency. Finally, we envision extending the MetaMolGen framework to broader tasks such as reaction pathway prediction, multi-step synthesis planning, and macromolecular design involving proteins and peptides.

\section*{Acknowledgment}
This work was supported by the National Natural Science Foundation of China [61773020] and the Graduate Innovation Project of National University of Defense Technology [XJQY2024065]. The authors would like to express their sincere gratitude to all the referees for their careful reading and insightful suggestions.

\clearpage
\onecolumn
\appendix

\section*{Appendix A: Benchmark and Baseline Details}
\label{app:data}

\subsection{Benchmark Datasets}
\label{app:benchmark}

MetaMolGen is evaluated on four widely used molecular datasets: ChEMBL, QM9, ZINC, and MOSES. These datasets provide a diverse range of chemical structures and physicochemical properties, allowing for a comprehensive evaluation of few-shot molecular generation capabilities.

ChEMBL \cite{Gaulton2012chembl} is used as the meta-training source dataset. It contains a large collection of bioactive, drug-like molecules curated from the scientific literature. We select a filtered subset of 100,000 molecules with molecular weight below 500 Daltons and QED scores above 0.5. Canonical SMILES strings are standardized using RDKit, and molecular graphs are extracted to serve as the input to graph- and sequence-based encoders. This dataset is used to construct meta-training tasks during episodic learning.

QM9 \cite{Ramakrishnan2014} is a dataset of approximately 134,000 small organic molecules with up to nine heavy atoms. Each molecule is annotated with a variety of quantum chemical properties. We sample a subset of 30,000 molecules to construct low-resource target tasks. Few-shot tasks are created by drawing random samples of 1,000 molecules as support sets, and 500 molecules as query sets. All molecules are preprocessed to remove invalid SMILES, and descriptors such as logP and HOMO-LUMO gap are normalized per-task.

ZINC \cite{Irwin2012} is a database of commercially available drug-like compounds. From the “clean leads” subset, we select between 10,000 and 50,000 molecules to simulate diverse chemical domains. Molecules are clustered based on Morgan fingerprint similarity (radius 2, length 2048), and structurally diverse scaffolds are used to create disjoint few-shot tasks. Tasks are constructed to assess how well models can generalize to previously unseen molecular motifs.

MOSES \cite{Polykovskiy2020} is a benchmark framework for molecule generation based on a curated subset of ZINC molecules. We use 60,000 molecules and follow the MOSES standard split for evaluation. This dataset provides strong baselines and standardized metrics, including validity, novelty, uniqueness, and scaffold diversity. For each MOSES task, 100–500 molecules are sampled as the support set, with 1,000 query molecules used for evaluation.

For all datasets, we ensure that SMILES are canonicalized, molecules are sanitized using RDKit, and invalid or charged species are removed. All features are normalized across datasets to reduce covariate shift. Summary statistics, such as average molecular length and number of unique scaffolds, are provided in Table~\ref{tab:benchmark-stats}.

\begin{table}[htbp]
\centering
\caption{Benchmark dataset statistics after preprocessing.}
\label{tab:benchmark-stats}
\begin{tabular}{lccc}
\toprule
Dataset & \# Molecules & Avg. SMILES Length & Unique Scaffolds \\
\midrule
ChEMBL & 100,000 & 27.4 & 15,320 \\
QM9 & 30,000 & 18.6 & 3,802 \\
ZINC & 50,000 & 24.1 & 7,632 \\
MOSES & 60,000 & 25.3 & 9,105 \\
\bottomrule
\end{tabular}
\end{table}

\vspace{1.5em}

\subsection{Compared Baselines}
\label{app:baselines}

We compare MetaMolGen against a diverse set of molecular generative baselines, covering both graph- and sequence-based architectures. All baseline models are either re-implemented or adapted from publicly available codebases. For fair comparison, all baselines are fine-tuned under the same few-shot protocol and evaluated using the same metrics and datasets.

MolGAN \cite{De Cao} is a graph-based implicit generative model that combines generative adversarial training with reinforcement learning for molecule optimization. We adopt the official implementation and fine-tune the generator on task-specific support sets using a reduced adversarial loss. The generator and critic use hidden dimensions of 128, with LeakyReLU activations. A fixed latent noise vector of length 56 is sampled per batch. Property-based reward signals are reweighted to emphasize validity and novelty in low-data settings.

GraphVAE \cite{Kusner2017GrammarVAE} is a variational autoencoder designed for small molecular graphs. The encoder-decoder pair is composed of graph convolutional networks (GCNs) and a fully connected projection to a 64-dimensional latent space. We pretrain GraphVAE on ChEMBL and fine-tune only the decoder parameters on few-shot tasks using a reconstruction loss. Maximum graph size is set to 38 atoms, and early stopping is applied based on query set loss.

CharRNN \cite{Segler} is a character-level recurrent model trained to autoregressively generate SMILES strings. We use a two-layer GRU with 512 hidden units, trained using teacher forcing with a ratio of 0.8. The model is pretrained on ChEMBL and fine-tuned on support sets for 20 epochs using a learning rate of 0.0005. During inference, beam search decoding with width 5 is applied to improve diversity.

TransformerVAE \cite{Lim J} is a transformer-based variational autoencoder that encodes SMILES sequences and decodes them via masked attention layers. The encoder consists of four transformer blocks with 8 attention heads and 256-dimensional hidden states. We use KL annealing and cyclical learning rate scheduling as suggested in the original implementation. In the few-shot setting, we perform 5 inner-loop adaptation steps on each task's support set.

All baselines are fine-tuned within the same meta-task format. For each task, only the support set is used for fine-tuning; the query set is held out for evaluation. Models are trained using Adam with weight decay 1e-6 and batch size 32 unless otherwise noted. We ensure that every method is exposed to the same number of training episodes and molecules for consistency.

We evaluate all models using five standardized metrics: validity (percentage of chemically valid outputs), uniqueness (non-duplicated outputs), novelty (not present in training), scaffold similarity (Tanimoto similarity of Bemis-Murcko scaffolds), and property compliance (e.g., QED or logP). Metrics are computed on 1,000 molecules sampled from each model per task using RDKit.

\section*{Appendix B: Theoretical Derivations}
\label{app:proofs}

This appendix provides full theoretical derivations that support the results presented in Section~\ref{SE5}. We begin by introducing two key lemmas used throughout the analysis: the Descent Lemma and a task encoder stability lemma. These results serve as foundational tools in proving the main convergence and generalization theorems.

For clarity, the primary theorems are restated below as Theorem 1 through Theorem 7, while preserving their original numbering in the main text (e.g., Theorem~\ref{TH1}–\ref{TH7}).

\begin{lemma}[Descent Lemma \cite{beck2017first}]
	If \( L_T(\theta) \) is \( L \)-smooth, then for any learning rate \( \alpha > 0 \), the loss after a gradient update satisfies:
	\begin{displaymath}
		L_T(\theta') \leq L_T(\theta) - \alpha \|\nabla L_T(\theta)\|^2 + \frac{L\alpha^2}{2} \|\nabla L_T(\theta)\|^2.
	\end{displaymath}
\end{lemma}

\begin{lemma}[Stability of Task Encoder \cite{bousquet2002stability}]
	Assume the task encoder $r_{\text{task}}$ in MetaMolGen is Lipschitz smooth with constant $L_r$, i.e., for any two datasets $D_1$ and $D_2$:
	\begin{displaymath}
		\| r_{\text{task}}(D_1) - r_{\text{task}}(D_2) \| \leq L_r \| D_1 - D_2 \|.
	\end{displaymath}
\end{lemma}

\vspace{1em}
\noindent\textbf{Theorem 1 (Convergence of Training) [Theorem~\ref{TH1} in main text].}
Let $L_T(\theta)$ be an $L$-smooth loss function. If the learning rate $\alpha$ satisfies
$0 < \alpha \leq \frac{2}{L},$
then the sequence of losses \( \{L_T(\theta_k)\} \) is monotonically decreasing, where

$$\theta_{k+1} = \theta_k - \alpha \nabla_{\theta} L_T(\theta_k).$$

\textit{Proof.} Since \(L_T(\theta)\) is assumed to be \(L\)-smooth, the Descent Lemma \cite{beck2017first} applies:
\begin{align}
    L_T(\theta_{k+1}) &\leq L_T(\theta_k) - \alpha \|\nabla L_T(\theta_k)\|^2 + \frac{L\alpha^2}{2} \|\nabla L_T(\theta_k)\|^2 \\
    &\leq L_T(\theta_k) - \left(\alpha - \frac{L\alpha^2}{2}\right)\|\nabla L_T(\theta_k)\|^2.
\end{align}

To ensure descent at each iteration:
\begin{align}
\alpha\left(1 - \frac{L\alpha}{2}\right) \geq 0 \quad \Rightarrow \quad \alpha \leq \frac{2}{L}.
\end{align}
Thus, loss is non-increasing under the condition. \hfill$\blacksquare$

\vspace{1em}
\noindent\textbf{Theorem 2 (Gradient Variance Reduction via Normalization) [Theorem~\ref{TH2}].}
Let $X \in \mathbb{R}^d$ be the input molecular descriptors, $X' = \frac{X - \hat{\mu}}{\hat{\sigma} + \epsilon}$ denote their normalized version. $\theta$ and $\theta'$ are model parameters obtained by training on $X$ and $X'$ , respectively. Then,
$$\operatorname{Var}(\nabla_\theta L_T(\theta')) \leq \operatorname{Var}(\nabla_\theta L_T(\theta)).$$

\textit{Proof.} The gradient variance is:
\begin{align}
\operatorname{Var}(\nabla_\theta L_T) = \mathbb{E}[\|\nabla_\theta L_T\|^2] - \|\mathbb{E}[\nabla_\theta L_T]\|^2.
\end{align}
Assume \(L_T\) is twice differentiable. If the Hessian has eigenvalues \(\lambda_j\), and the input variance per feature is \(\sigma_j^2\), then:
\begin{align}
\mathbb{E}[\|\nabla_\theta L_T\|^2] \approx \sum_{j=1}^d \lambda_j \sigma_j^2.
\end{align}
After normalization: \(\sigma_j^2 \approx 1\). If \(\sigma_j^2 > 1\) before, normalization reduces the sum, and thus gradient variance.
\hfill$\blacksquare$

\vspace{1em}
\noindent\textbf{Theorem 3 (Improved Conditioning and Accelerated Convergence) [Theorem~\ref{TH3}].}
Let $L_T(\theta)$ be twice differentiable, with Hessian $H = \nabla^2_\theta L_T(\theta)$ and condition number $\kappa(H) = \lambda_{\max}(H) / \lambda_{\min}(H)$. Consider normalized inputs $$X' = (X - \hat{\mu}) / (\hat{\sigma} + \epsilon),$$ and let $L_T'(\theta)$ be the corresponding loss with Hessian $H' = \nabla^2_\theta L_T'(\theta)$. Then, $$\kappa(H') \ll \kappa(H),$$ and the iteration complexity of first-order optimization improves from $O\left(\kappa(H) \log \tfrac{1}{\epsilon}\right)$ to $O\left(\log \tfrac{1}{\epsilon}\right)$.

\textit{Proof.}
Let \( x \in \mathbb{R}^d \) denote the input features and define the empirical loss as
\begin{align}
L_T(\theta) &= \frac{1}{n} \sum_{i=1}^n \ell(f_\theta(x^{(i)}), y^{(i)}),
\end{align}
where \( f_\theta \colon \mathbb{R}^d \to \mathbb{R} \) is a differentiable model, and \( \ell \) is a convex, twice-differentiable loss function. The Hessian is given by
\begin{align}
H &= \nabla^2_\theta L_T(\theta) = \frac{1}{n} \sum_{i=1}^n \nabla^2_\theta \ell(f_\theta(x^{(i)}), y^{(i)}).
\end{align}

Let the input second-moment matrix be
\begin{align}
\Sigma_x &= \mathbb{E}[x x^\top].
\end{align}
If the eigenvalue ratio satisfies
\begin{align}
\frac{\lambda_{\max}(\Sigma_x)}{\lambda_{\min}(\Sigma_x)} &\gg 1,
\end{align}
then there exists a unit vector \( u \in \mathbb{R}^d \) such that
\begin{align}
\operatorname{Var}(\langle x, u \rangle) &= u^\top \Sigma_x u \gg v^\top \Sigma_x v, \quad \forall v \perp u.
\end{align}
This anisotropy propagates through the model's Jacobian \( J_\theta(x) := \nabla_\theta f_\theta(x) \), leading to a dominant Hessian contribution
\begin{align}
H &\approx \mathbb{E}\left[ J_\theta(x)^\top \nabla^2_f \ell(f_\theta(x), y) J_\theta(x) \right].
\end{align}
When the feature scales vary across dimensions, the rows of \( J_\theta(x) \) inherit these scales, which results in unbalanced eigenvalues and hence
\begin{align}
\kappa(H) &= \frac{\lambda_{\max}(H)}{\lambda_{\min}(H)} \gg 1.
\end{align}

Feature normalization is applied as
\begin{align}
x_i' &= \frac{x_i - \mu_i}{\sigma_i}, \quad \mu_i = \mathbb{E}[x_i], \quad \sigma_i^2 = \operatorname{Var}(x_i),
\end{align}
so that the transformed inputs satisfy
\begin{align}
\mathbb{E}[x' x'^\top] &= I_d.
\end{align}
As a result, the expected Jacobian structure becomes more uniform:
\begin{align}
\mathbb{E}[J_\theta(x')^\top J_\theta(x')] &\approx c \cdot I,
\end{align}
for some constant \( c > 0 \), implying that the model responds equally to perturbations in all directions of parameter space. This leads to
\begin{align}
\kappa(H') &\ll \kappa(H), \quad \text{and ideally} \quad \kappa(H') \approx 1.
\end{align}

The iteration complexity of gradient descent for minimizing a strongly convex function with condition number \(\kappa\) satisfies
\begin{align}
T &= O\left(\kappa(H) \log \tfrac{1}{\epsilon} \right),
\end{align}
which improves after normalization to
\begin{align}
T &= O\left( \log \tfrac{1}{\epsilon} \right),
\end{align}
completing the proof.
\hfill\(\blacksquare\)

\vspace{1em}
\noindent\textbf{Theorem 4 (Variance Reduction and Generalization Improvement) [Theorem~\ref{TH4}].}
Let $X \in \mathbb{R}^d$ be molecular descriptors, and $\mathcal{E}_{\text{meta, raw}}$, $\mathcal{E}_{\text{meta, standardized}}$ be the meta-generalization errors under training on $X$ and its normalized form $X' = (X - \hat{\mu}) / (\hat{\sigma} + \epsilon)$, respectively. Then,
$$
\mathbb{E}[\mathcal{E}_{\text{meta, standardized}}] < \mathbb{E}[\mathcal{E}_{\text{meta, raw}}].
$$

\textit{Proof.}
According to the PAC-Bayes generalization theory \cite{mcallester1999some}, for a learned posterior distribution \( Q \) over hypotheses and a prior \( P \), the expected generalization error satisfies
\begin{align}
\mathbb{E}[\mathcal{E}_{\text{meta}}] 
&\leq \mathbb{E}[\mathcal{L}_{\text{train}}] + \mathcal{C}(Q \| P, N), \\
\mathcal{C}(Q \| P, N) 
&= O\left( \frac{D_{\mathrm{KL}}(Q \| P) + \log(1/\delta)}{N} \right),
\end{align}
where \( N \) is the number of meta-training episodes and \( \delta \in (0,1) \) is the failure probability.

If \( Q \) is modeled as a Gaussian centered at the empirical minimizer, then the KL divergence \( D_{\mathrm{KL}}(Q \| P) \) scales with the empirical parameter variance:
\begin{align}
D_{\mathrm{KL}}(Q \| P) \propto \frac{1}{\sigma^2},
\end{align}
where \( \sigma^2 \) denotes the empirical loss variance over episodes.

For first-order meta-learning algorithms such as Reptile, applied over support/query splits of sizes \( N_s, N_q \), this yields a generalization bound of the form:
\begin{align}
\mathbb{E}[\mathcal{E}_{\text{meta}}] 
&\leq \mathbb{E}[\mathcal{L}_{\text{train}}] 
+ O\left( \frac{\sigma^2}{N_s} + \frac{\sigma^2}{N_q} \right).
\end{align}

Let \( x \in \mathbb{R}^d \) denote the input features, and let normalization transform each feature as
\begin{align}
x'_i = \frac{x_i - \mu_i}{\sigma_i}.
\end{align}
This transformation enforces unit variance in each dimension:
\begin{align}
\mathbb{E}[{x'}_i^2] = 1, \quad \forall i.
\end{align}
As the loss function \(\ell(f_\theta(x), y)\) is typically Lipschitz-continuous in \( x \), i.e.,
\begin{align}
|\ell(f_\theta(x), y) - \ell(f_\theta(x'), y)| \leq L \|x - x'\|,
\end{align}
the reduction of feature variance translates into smaller loss fluctuations across episodes, implying that
\begin{align}
\sigma^2_{\text{standardized}} < \sigma^2_{\text{raw}}.
\end{align}

Consequently, the generalization bound becomes tighter under normalization:
\begin{align}
\mathbb{E}[\mathcal{E}_{\text{meta, standardized}}] < \mathbb{E}[\mathcal{E}_{\text{meta, raw}}].
\end{align}
Additionally, lower loss variance implies higher task stability and smoother adaptation in gradient-based meta-updates.
\hfill\(\blacksquare\)

\vspace{1em}
\noindent\textbf{Theorem 5 (Unbiasedness of Stochastic Gradient) [Theorem~\ref{TH5}].}
Let $\mathcal{D}_T = \{(x_i, z_i, y_i)\}_{i=1}^{N_T}$ be the dataset for task $T$, where $x_i$ is the molecular input, $z_i$ is the task-specific context, and $y_i$ is the corresponding label. Let $\mathcal{B} \subset \mathcal{D}_T$ be a mini-batch sampled uniformly. The stochastic gradient estimator of $L_T(\theta)$ is defined as
$$
\nabla_{\text{est}} L_T(\theta) = \frac{1}{|\mathcal{B}|} \sum_{(x_i, z_i, y_i) \in \mathcal{B}} \nabla_\theta \mathcal{L}(f_\theta(x_i, z_i), y_i),
$$
where $f_\theta$ is the model and $\mathcal{L}$ is the sample-wise loss. Then the estimator is unbiased:
$$
\mathbb{E}_{\mathcal{B}}[\nabla_{\text{est}} L_T(\theta)] = \nabla_\theta L_T(\theta).
$$

\textit{Proof.}
Let the per-sample gradient be denoted as
\begin{align}
g_i(\theta) := \nabla_\theta \mathcal{L}(f_\theta(x_i, z_i), y_i).
\end{align}
Then the full-batch gradient is expressed as
\begin{align}
\nabla_\theta L_T(\theta) = \frac{1}{N_T} \sum_{i=1}^{N_T} g_i(\theta).
\end{align}

Let \(\mathcal{B} = \{i_1, i_2, \dots, i_B\}\) be a set of indices sampled uniformly without replacement from \(\{1, 2, \dots, N_T\}\). The stochastic gradient estimator becomes
\begin{align}
\mathbb{E}_{\mathcal{B}}[\nabla_{\text{est}} L_T(\theta)] 
&= \mathbb{E}_{\mathcal{B}}\left[ \frac{1}{B} \sum_{j=1}^{B} g_{i_j}(\theta) \right] \\
&= \frac{1}{B} \sum_{j=1}^{B} \mathbb{E}_{i_j}[g_{i_j}(\theta)] \\
&= \frac{1}{B} \sum_{j=1}^{B} \left( \frac{1}{N_T} \sum_{i=1}^{N_T} g_i(\theta) \right).
\end{align}

This simplifies to
\begin{align}
\frac{1}{B} \cdot B \cdot \frac{1}{N_T} \sum_{i=1}^{N_T} g_i(\theta) = \nabla_\theta L_T(\theta).
\end{align}
Hence, the stochastic gradient estimator is an unbiased estimator of the true gradient.
\hfill\(\blacksquare\)

\vspace{1em}
\noindent\textbf{Theorem 6 (Convergence under Stochastic Updates) [Theorem~\ref{TH6}].}
Let $L_T(\theta)$ be an $L$-smooth loss function, and the stochastic gradient estimator is unbiased with bounded variance $\sigma^2$. The parameter after $k$ iterations of SGD is denoted by $\theta_k$, and $\theta^*$ be the minimizer of $L_T(\theta)$. Then the expected suboptimality satisfies:
$$
\mathbb{E}[L_T(\theta_k)] - L_T(\theta^*) = O\left( \frac{L}{k} \right).
$$

\textit{Proof.}
By the smoothness of \(L_T\), we have:
\begin{align}
L_T(\theta_{k+1}) 
&\leq L_T(\theta_k) + \langle \nabla L_T(\theta_k), \theta_{k+1} - \theta_k \rangle + \frac{L}{2} \|\theta_{k+1} - \theta_k\|^2.
\end{align}

Let the update rule be:
\begin{align}
\theta_{k+1} &= \theta_k - \alpha \nabla_{\text{est}} L_T(\theta_k),
\end{align}
then plugging into the inequality:
\begin{align}
\mathbb{E}[L_T(\theta_{k+1})] 
&\leq \mathbb{E}[L_T(\theta_k)] - \alpha \mathbb{E}[\|\nabla L_T(\theta_k)\|^2] + \frac{L \alpha^2}{2} \mathbb{E}[\|\nabla_{\text{est}} L_T(\theta_k)\|^2].
\end{align}

Expanding the second moment via bias-variance decomposition:
\begin{align}
\mathbb{E}[\|\nabla_{\text{est}} L_T(\theta_k)\|^2] 
&= \|\nabla L_T(\theta_k)\|^2 + \mathbb{E}[\|\nabla_{\text{est}} L_T(\theta_k) - \nabla L_T(\theta_k)\|^2] \\
&\leq \|\nabla L_T(\theta_k)\|^2 + \sigma^2.
\end{align}

Substituting back:
\begin{align}
\mathbb{E}[L_T(\theta_{k+1})] 
&\leq \mathbb{E}[L_T(\theta_k)] - \alpha \mathbb{E}[\|\nabla L_T(\theta_k)\|^2] + \frac{L \alpha^2}{2} \left( \mathbb{E}[\|\nabla L_T(\theta_k)\|^2] + \sigma^2 \right).
\end{align}

Grouping terms:
\begin{align}
\mathbb{E}[L_T(\theta_{k+1})] 
&\leq \mathbb{E}[L_T(\theta_k)] - \alpha\left(1 - \frac{L\alpha}{2} \right)\mathbb{E}[\|\nabla L_T(\theta_k)\|^2] + \frac{L\alpha^2}{2} \sigma^2.
\end{align}

Let \(\alpha \leq \frac{1}{L}\), then \(1 - \frac{L\alpha}{2} \geq \frac{1}{2}\), so we obtain:
\begin{align}
\mathbb{E}[L_T(\theta_{k+1})] 
&\leq \mathbb{E}[L_T(\theta_k)] - \frac{\alpha}{2} \mathbb{E}[\|\nabla L_T(\theta_k)\|^2] + \frac{L\alpha^2}{2} \sigma^2.
\end{align}

Telescoping the inequality over \(k\) steps and using \(L_T(\theta_k) \geq L_T(\theta^*)\), we obtain:
\begin{align}
\frac{1}{k} \sum_{j=1}^k \mathbb{E}[\|\nabla L_T(\theta_j)\|^2] 
&\leq O\left(\frac{1}{k}\right),
\end{align}
which implies the sublinear convergence in function value:
\begin{align}
\mathbb{E}[L_T(\theta_k)] - L_T(\theta^*) &= O\left( \frac{1}{k} \right).
\end{align}
\hfill\(\blacksquare\)

\vspace{1em}
\noindent\textbf{Theorem 7 (Generalization and Error Bound) [Theorem~\ref{TH7}].}  
Let \(\theta_k\) be the model parameters after \(k\) updates, and the optimal parameters minimizing expected loss is denoted by $\theta^*$. Then the total expected error satisfies:
\begin{align*}
\mathbb{E}_{T \sim p(T)}[L_T(\theta_k)] - L_T(\theta^*) 
&\leq O\left( \frac{L}{k} \right) + O\left( \sqrt{ \frac{D_{\text{KL}}(Q \| P)}{N} } \right) \\
&\quad + O\left( \frac{\sigma}{\sqrt{B}} \right) + O(\epsilon_{\text{approx}}),
\end{align*}
where the four terms represent: 
$O\left( \frac{L}{k} \right)$ for optimization error, 
$O\left( \sqrt{ \frac{D_{\text{KL}}(Q \| P)}{N} } \right)$ for generalization error, 
$O\left( \frac{\sigma}{\sqrt{B}} \right)$ for gradient noise, 
and $O(\epsilon_{\text{approx}})$ for approximation error.

\textit{Proof.}  
The total expected loss gap \(\mathbb{E}_{T}[L_T(\theta_k)] - L_T(\theta^*)\) can be decomposed into four sources of error. First, the optimization error arises from the convergence behavior of stochastic gradient descent. From Theorem~\ref{TH6}, under the assumption that the loss \(L_T(\theta)\) is \(L\)-smooth and gradients are unbiased with bounded variance, the optimization error decays as \(O\left(\frac{L}{k}\right)\) after \(k\) updates with the appropriate step size \(\alpha = O(1/L)\).

Second, the generalization error captures the discrepancy between the empirical and population risk. Applying the PAC-Bayes bound \cite{mcallester1999some}, we obtain:
\begin{align}
\mathbb{E}_{T \sim p(T)}[L_T(\theta_k)] \leq \mathbb{E}_{T \sim p(T)}[\hat{L}_T(\theta_k)] + O\left( \sqrt{ \frac{D_{\text{KL}}(Q \| P)}{N} } \right),
\end{align}
where \(Q\) is the posterior distribution over parameters induced by training, \(P\) is the prior (often task-independent), and \(N\) is the number of training tasks. The bound tightens as more tasks are observed.

Third, the stochastic gradient estimation noise originates from using mini-batch updates. From Theorem~\ref{TH5}, the gradient estimator is unbiased, but its variance contributes additional error. Under batch size \(B\), the variance of the mini-batch gradient scales inversely with \(B\), yielding an error term \(O\left(\frac{\sigma}{\sqrt{B}}\right)\), where \(\sigma^2\) is the upper bound on per-task gradient variance.

Finally, the approximation error reflects the expressive limitation of the model class \(\mathcal{F}_\theta\). Even if optimization and generalization were perfect, the learned model might still underfit complex molecular functions. We denote this irreducible gap by \(\epsilon_{\text{approx}}\), corresponding to the best achievable loss within the hypothesis class.

Combining all four components, we obtain the final bound:
\begin{align}
\mathbb{E}_{T \sim p(T)}[L_T(\theta_k)] - L_T(\theta^*) \leq O\left( \frac{L}{k} \right) + O\left( \sqrt{ \frac{D_{\text{KL}}(Q \| P)}{N} } \right) + O\left( \frac{\sigma}{\sqrt{B}} \right) + O(\epsilon_{\text{approx}}).
\end{align}
\hfill$\blacksquare$

\section*{Appendix C: Supplementary Results}
\label{app:supp}

\begin{figure*}[htbp]
    \centering
    \begin{subfigure}[t]{0.45\linewidth}
        \centering
        \includegraphics[width=\linewidth]{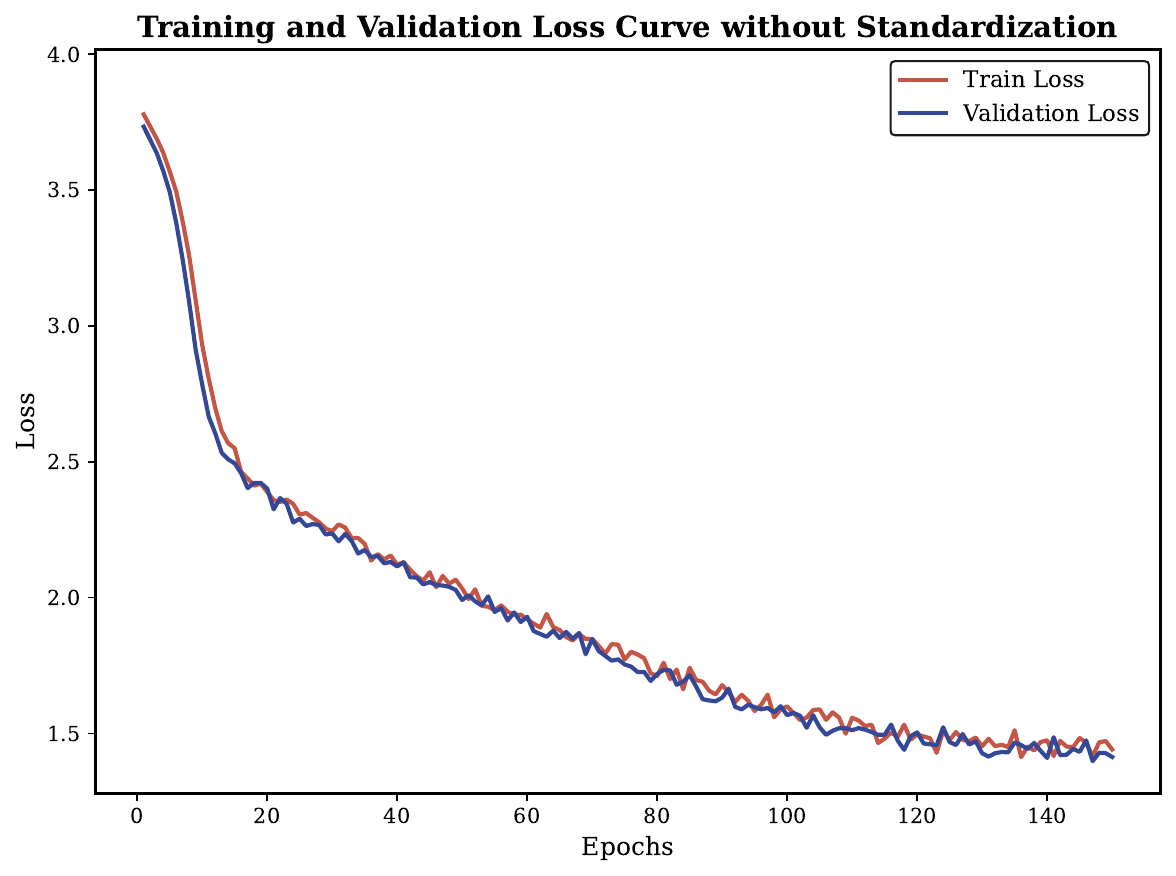}
        \caption{Without feature standardization}
    \end{subfigure}
    \hfill
    \begin{subfigure}[t]{0.45\linewidth}
        \centering
        \includegraphics[width=\linewidth]{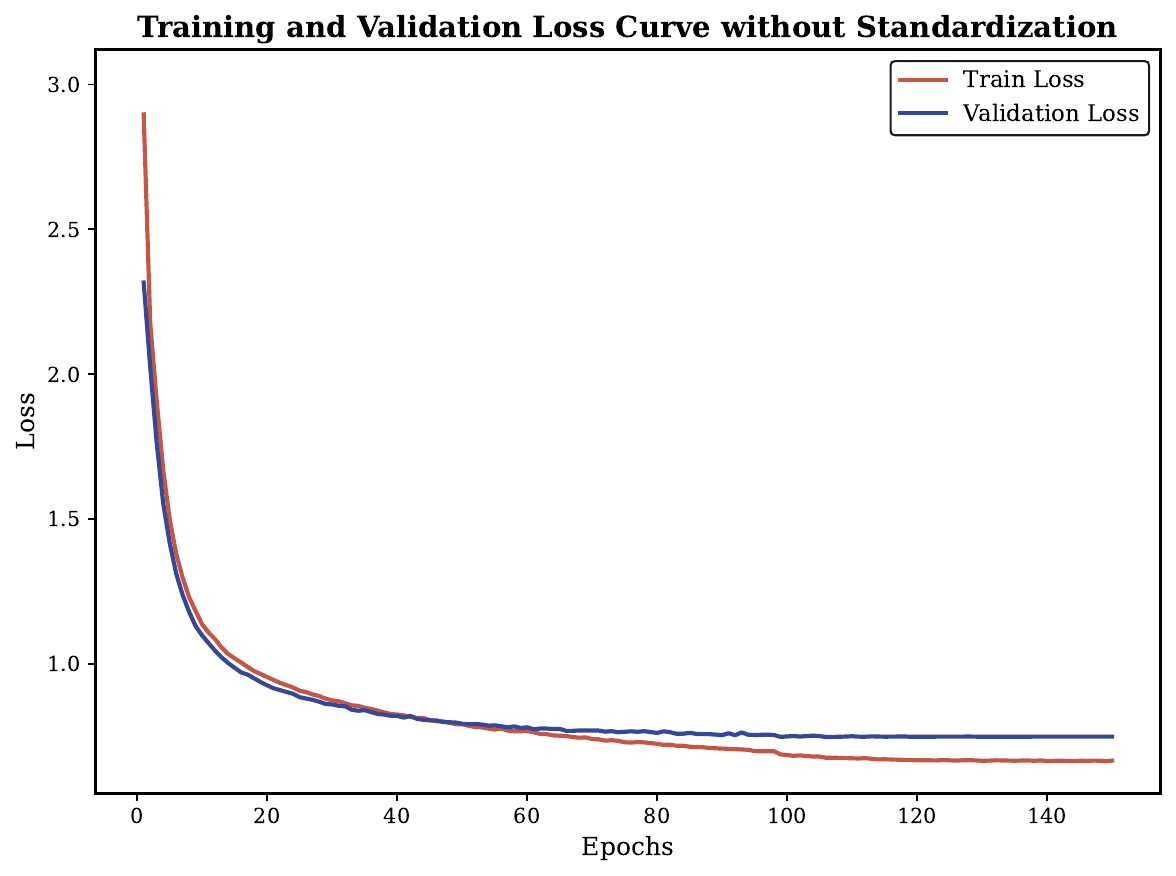}
        \caption{With feature standardization}
    \end{subfigure}
    \caption{Training loss trajectories with and without feature standardization. Feature standardization leads to smoother convergence and reduced variance during meta-adaptation.}
    \label{fig:meta_loss_comparison}
\end{figure*}

This section presents supplementary results that are omitted from the main text due to space limitations. These include:

\begin{enumerate}
    \item Visualization of the effect of standardization on training dynamics through loss trajectories (Figure~\ref{fig:meta_loss_comparison}).
    
    \item Full metric breakdowns across training sizes for few-shot molecular generation (Table~\ref{tab:fewshot_valid}).

    \item Detailed results from the ablation study on the effect of the standardization layer (Table~\ref{tab:ablation_study}).
\end{enumerate}

\vspace{1em}

\textbf{Effect of Feature Standardization on Meta-Training Loss.}  
Figure~\ref{fig:meta_loss_comparison} shows the training loss trajectories of MetaMolGen with and without feature standardization. Feature standardization smooths the optimization landscape, leading to more stable convergence and lower variance during meta-updates, as discussed in Section~3.2 of the main text.

\vspace{1em}

\textbf{Few-shot Performance Metrics.}  
Table~\ref{tab:fewshot_valid} shows the complete comparison of MetaMolGen, RNN, and MolGPT under training set sizes ranging from 1,000 to 10,000 samples. MetaMolGen consistently achieves the best scores in validity, drug-likeness, and overall generation quality, with particularly strong performance under low-resource conditions.

\begin{table*}[h]
\centering
\caption{Comparison of model performance across training sizes and metrics. Gray cells indicate the best value per column (i.e., per training size).}
\small
\resizebox{\textwidth}{!}{
\begin{tabular}{ll|cccccccccc}
\toprule
\textbf{Metric} & \textbf{Model} & \textbf{1000} & \textbf{2000} & \textbf{3000} & \textbf{4000} & \textbf{5000} & \textbf{6000} & \textbf{7000} & \textbf{8000} & \textbf{9000} & \textbf{10000} \\
\midrule
\multirow{3}{*}{Valid (\%)} 
  & RNN        & 34.80 & 44.20 & 31.80 & 48.60 & 49.00 & 49.00 & 58.40 & 61.40 & 47.80 & 58.60 \\
  & MolGPT     &  1.60 & 6.00 & 8.00 & 16.20 & 25.20 & 20.80 & 2.80 & 20.60 & 10.20 & 11.80 \\
  & MetaMolGen & \cellcolor{gray!25}43.60 & \cellcolor{gray!25}60.20 & \cellcolor{gray!25}66.00 & \cellcolor{gray!25}67.80 & \cellcolor{gray!25}76.00 & \cellcolor{gray!25}78.20 & \cellcolor{gray!25}80.80 & \cellcolor{gray!25}82.80 & \cellcolor{gray!25}82.40 & \cellcolor{gray!25}83.60 \\
\midrule
\multirow{3}{*}{Diversity}
  & RNN        & 0.8619 & 0.8707 & 0.8671 & 0.8536 & \cellcolor{gray!25}0.8746 & \cellcolor{gray!25}0.8718 & 0.8680 & \cellcolor{gray!25}0.8852 & 0.8593 & 0.8681 \\
  & MolGPT     & \cellcolor{gray!25}0.8914 & 0.8481 & \cellcolor{gray!25}0.8819 & \cellcolor{gray!25}0.8848 & 0.8634 & 0.8649 & \cellcolor{gray!25}0.8760 & 0.8747 & \cellcolor{gray!25}0.8905 & \cellcolor{gray!25}0.8735 \\
  & MetaMolGen & 0.8285 & \cellcolor{gray!25}0.8370 & 0.8374 & 0.8367 & 0.8374 & 0.8387 & 0.8374 & 0.8401 & 0.8383 & 0.8415 \\
\midrule
\multirow{3}{*}{Druglikeness}
  & RNN        & 0.5278 & 0.5293 & 0.5961 & 0.5343 & 0.5487 & 0.5226 & 0.5007 & 0.6143 & 0.5653 & 0.5426 \\
  & MolGPT     & 0.5129 & 0.4392 & 0.5978 & 0.6184 & 0.5162 & 0.6205 & 0.6483 & 0.5961 & 0.5102 & 0.5951 \\
  & MetaMolGen & \cellcolor{gray!25}0.8133 & \cellcolor{gray!25}0.8128 & \cellcolor{gray!25}0.8215 & \cellcolor{gray!25}0.8195 & \cellcolor{gray!25}0.8160 & \cellcolor{gray!25}0.8156 & \cellcolor{gray!25}0.8193 & \cellcolor{gray!25} 0.8165 & \cellcolor{gray!25} 0.8176 & \cellcolor{gray!25}0.8090 \\
\midrule
\multirow{3}{*}{Overall Score}
  & RNN        & 0.5262 & 0.5108 & 0.6006 & 0.5349 & 0.5176 & 0.5059 & 0.4883 & 0.5841 & 0.5396 & 0.5209 \\
  & MolGPT     & 0.4541 & 0.3640 & 0.5053 & 0.5561 & 0.4831 & 0.5639 & 0.5263 & 0.5533 & 0.4078 & 0.5111 \\
  & MetaMolGen & \cellcolor{gray!25}0.6857 & \cellcolor{gray!25}0.6859 & \cellcolor{gray!25}0.6966 & \cellcolor{gray!25}0.6946 & \cellcolor{gray!25}0.6917 & \cellcolor{gray!25}0.7097 & \cellcolor{gray!25}0.7300 & \cellcolor{gray!25}0.7447 & \cellcolor{gray!25}0.7405 & \cellcolor{gray!25}0.7398 \\
\bottomrule
\end{tabular}
}
\label{tab:fewshot_valid}
\end{table*}

\vspace{1em}

\textbf{Ablation Study Results.}  
Table~\ref{tab:ablation_study} provides a complete breakdown of the ablation results comparing model performance with and without the standardization layer across various training sizes. These values correspond to the visual trends discussed in the main text Figure~\ref{fig:ablation_combined}.

\begin{table*}[h]
\centering
\caption{Ablation study: Evaluation of validity, novelty, diversity, and CGSR with and without standardization. Gray cells indicate better performance.}
\small
\resizebox{\textwidth}{!}{
\begin{tabular}{c|cc|cc|cc|cc|cc}
\toprule
\multirow{2}{*}{\textbf{Training Data Size}} & \multicolumn{2}{c|}{\textbf{Validity (\%)}} & \multicolumn{2}{c|}{\textbf{Novelty (\%)}} & \multicolumn{2}{c|}{\textbf{Diversity}} & \multicolumn{2}{c|}{\textbf{CGSR (NHA) (\%)}} & \multicolumn{2}{c}{\textbf{CGSR (NHD) (\%)}} \\
\cline{2-11}
& Without & With & Without & With & Without & With & Without & With & Without & With \\
\midrule
60,000 & 89.25 & \cellcolor{gray!25}92.80 & 89.25 & \cellcolor{gray!25}92.80 & \cellcolor{gray!25}0.8500 & 0.8485 & 74.12 & 71.86 & 93.42 & \cellcolor{gray!25}95.67 \\
30,000 & 76.20 & \cellcolor{gray!25}88.20 & 76.20 & \cellcolor{gray!25}88.20 & \cellcolor{gray!25}0.8492 & 0.8451 & \cellcolor{gray!25}80.07 & 74.78 & 94.26 & 94.03 \\
10,000 & \cellcolor{gray!25}83.80 & 83.60 & \cellcolor{gray!25}83.80 & 83.60 & \cellcolor{gray!25}0.8462 & 0.8426 & \cellcolor{gray!25}81.26 & 79.71 & \cellcolor{gray!25}95.55 & 95.13 \\
6,000  & 73.80 & \cellcolor{gray!25}78.20 & 73.80 & \cellcolor{gray!25}78.20 & 0.8476 & \cellcolor{gray!25}0.8532 & 72.16 & \cellcolor{gray!25}75.26 & 93.24 & \cellcolor{gray!25}92.65 \\
3,000  & 62.20 & \cellcolor{gray!25}66.00 & 62.20 & \cellcolor{gray!25}66.00 & 0.8434 & \cellcolor{gray!25}0.8503 & 82.32 & 81.48 & 92.93 & 91.58 \\
1,000  & 36.80 & \cellcolor{gray!25}43.60 & 36.80 & \cellcolor{gray!25}43.60 & 0.8332 & \cellcolor{gray!25}0.8476 & 80.98 & \cellcolor{gray!25}82.13 & 91.85 & \cellcolor{gray!25}94.67 \\
\bottomrule
\end{tabular}
}
\label{tab:ablation_study}
\end{table*}

\end{document}